\documentclass[sn-basic]{sn-jnl}

\usepackage{graphicx}%
\usepackage{multirow}%
\usepackage{amsmath}%
\usepackage{amssymb}%
\usepackage{amsfonts}%
\usepackage{mathrsfs}%
\usepackage[title]{appendix}%
\usepackage[table]{xcolor}%
\usepackage{textcomp}%
\usepackage{manyfoot}%
\usepackage{booktabs}%
\usepackage{algorithm}%
\usepackage{algorithmicx}%
\usepackage{algpseudocode}%
\usepackage{listings}%
\usepackage{longtable}%
\usepackage{pdflscape}%
\usepackage{natbib}%
\usepackage{array}%
\usepackage{tabularray}%

\begin{document}

\title[Article Title]{medicX-KG: A Knowledge Graph for Pharmacists' Drug Information Needs}


\author[1]{\fnm{Lizzy} \sur{Farrugia}}\email{lizzy.farrugia@um.edu.mt}

\author[2]{\fnm{Lilian M.} \sur{Azzopardi}}\email{lilian.azzopardi@um.edu.mt}
\author[1]{\fnm{Jeremy} \sur{Debattista}}\email{jerdebattista@gmail.com}
\author*[1]{\fnm{Charlie} \sur{Abela}}\email{charlie.abela@um.edu.mt}

\affil*[1]{\orgdiv{Department of Artificial Intelligence}, \orgname{University of Malta}, \orgaddress{\city{Msida}, \country{Malta}}}

\affil[2]{\orgdiv{Department of Pharmacy}, \orgname{University of Malta}, \orgaddress{\city{Msida}, \country{Malta}}}


\abstract{
The role of pharmacists is evolving from medicine dispensing to delivering comprehensive pharmaceutical services within multidisciplinary healthcare teams. Central to this shift is access to accurate, up-to-date medicinal product information supported by robust data integration. Leveraging artificial intelligence and semantic technologies, Knowledge Graphs (KGs) uncover hidden relationships and enable data-driven decision-making. This paper presents medicX-KG, a pharmacist-oriented knowledge graph supporting clinical and regulatory decisions. It forms the semantic layer of the broader \textit{medicX} platform, powering predictive and explainable pharmacy services. medicX-KG integrates data from three sources, including, the British National Formulary (BNF), DrugBank, and the Malta Medicines Authority (MMA) that addresses Malta’s regulatory landscape and combines European Medicines Agency alignment with partial UK supply dependence. The KG tackles the absence of a unified national drug repository, reducing pharmacists’ reliance on fragmented sources. Its design was informed by interviews with practising pharmacists to ensure real-world applicability. We detail the KG’s construction, including data extraction, ontology design, and semantic mapping. Evaluation demonstrates that medicX-KG effectively supports queries about drug availability, interactions, adverse reactions, and therapeutic classes. Limitations, including missing detailed dosage encoding and real-time updates, are discussed alongside directions for future enhancements.
}



\maketitle

\section{Introduction}\label{sec:introduction}

Pharmacy practice has evolved significantly from the traditional role of dispensing medicines to an integral function within multidisciplinary healthcare teams~\citep{azzopardi2020}. Pharmacists now routinely contribute to the optimisation of medication therapy, the monitoring of adverse drug reactions (ADRs), and clinical decision making~\citep{fip2023}. These responsibilities have become increasingly complex with the rise of polypharmacy, commonly defined as the simultaneous use of five or more medications~\citep{masnoon2017}, which substantially increases the risk of drug-drug interactions (DDI) and ADRs~\citep{zitnik2018, masumshah2021, kontsioti2022}.

As the frequency and complexity of DDIs increase, particularly in elderly and poly-medicated populations, access to accurate, timely, and locally relevant drug information has become essential. However, a systematic review by~\cite{tan2023} found that pharmacists primarily rely on reference texts such as the British National Formulary (BNF)~\citep{bnf2024} and subscription-based databases such as UpToDate~\citep{uptodate2024} and Micromedex~\citep{micromedex2024}. Although widely used (84.6\% and 69.2\% of pharmacists surveyed, respectively), these sources exhibit notable limitations, including concerns about credibility (50\%), insufficient or fragmented coverage (42.9\%), usability challenges (42.9\%) and high subscription costs (21.4\%). In addition, inconsistencies in DDI reporting across resources such as BNF, UpToDate, and Micromedex have been highlighted in recent comparative studies~\citep{kontsioti2022, shariff2022}, exposing pharmacists to potential uncertainty in clinical decisions.

The national regulatory contexts further complicate this landscape. The Malta Medicines Authority (MMA)~\citep{mma2024} is legally aligned with the European Medicines Agency (EMA)~\citep{ema2024} through Directive 2001/83/EC~\citep{santfournier2020, mma2020}. However, more 65\% of Malta's medicinal products continue to be sourced through UK regulatory and commercial channels~\citep{euamr2024}. Maltese pharmacists, therefore, navigate an information environment fragmented between EMA protocols and legacy UK references such as the BNF. This regulatory duality, compounded by the small size of the market, the absence of a unified national drug repository, and the language-specific labelling restrictions imposed by EU law, restricts the flexibility of product importation. Although Malta is bilingual (Maltese and English), EU requirements often require the availability of SmPCs or patient leaflets in Maltese, limiting options for importing products from non-English-speaking countries.

Such challenges are not isolated to Malta. Other EU markets, including Ireland and Cyprus, similarly face regulatory vulnerabilities and have secured post-Brexit derogations for medicinal product continuity~\citep{euamr2024}. This broader regional pattern underscores the need for infrastructures that not only bridge global and national regulatory frameworks but also offer semantic richness and operational relevance to pharmacists.

Knowledge graphs (KG) have shown considerable promise in the biomedical domain in integrating heterogeneous datasets, supporting semantic querying, and allowing inference over relationships such as therapeutic classes, contraindications, and DDI~\citep{himmelstein2017, percha2018}. Their ability to harmonise structurally diverse datasets into a coherent semantic layer makes them particularly well-suited to environments marked by jurisdictional overlap and terminological inconsistencies.

In this paper, we present \textit{medicX-KG}, a pharmacist-oriented biomedical knowledge graph embedded within the broader \textit{medicX} platform~\citep{farrugia2020, farrugia2023}, where 'X' denotes the focus of the system on drug-drug interactions. Our goal is to design a semantically enriched drug information infrastructure that addresses both the dual regulatory realities and the practical clinical needs of pharmacists operating in Malta and comparable jurisdictions. 

Although existing biomedical KGs such as Bio2RDF~\citep{belleau2008}, Hetionet~\citep{himmelstein2017}, DRKG~\citep{ioannidis2020}, PrimeKG~\citep{chandak2023}, and PharmaKG~\citep{asada2021} offer broad coverage of biomedical entities and relationships, they are not tailored to country-specific availability, labelling, or regulatory status. Furthermore, mainstream clinical databases such as DailyMed~\citep{dailymed2025}, RxNorm~\citep{le2024}, First DataBank~\citep{fdb2025}, UpToDate~\citep{uptodate2024}, and Micromedex~\citep{micromedex2024} tend to silo pharmacological content without enabling integrated reasoning across local regulatory dimensions.

The novelty of medicX-KG lies in its harmonised integration of local and international drug sources through an ontology-driven mapping framework. This includes structured mappings across the BNF~\citep{bnf2024}, the MMA product listings~\citep{mma2024}, and DrugBank~\citep{wishart2023}, an especially challenging task given the inconsistencies in product naming, availability, and regulatory labelling.

To ensure that KG aligns with real-world pharmacist needs, we conducted a focus group study involving professionals from the community, hospital, and regulatory pharmacy sectors. Thematic analysis of the interview data identified key pain points, including the dependence on subscription-based tools, the absence of product equivalence data, and fragmented search capabilities across available systems. These findings directly informed the semantic schema, entity relations, and query design of KG.

We evaluated medicX-KG through competency questions grounded in real-world pharmacy scenarios, focusing on use cases such as polypharmacy risk detection, therapeutic substitutions, ADR verification, and local product availability checks.

In summary, medicX-KG addresses the critical and underexplored need to enable pharmacists operating under constrained regulatory and linguistic environments to access, reason over, and act on reliable drug information. Its design reflects both technical innovation and domain-driven contextualisation, providing a transferable blueprint for locality-aware KG development in similarly underserved healthcare ecosystems.

\section{Related Work}
Recent years have seen a proliferation of biomedical knowledge graphs (BKGs) aimed at integrating heterogeneous data sources to support applications in drug discovery, genomic medicine, and pharmacovigilance. Although these graphs offer powerful tools for reasoning about drug-disease relationships, adverse events, and therapeutic mechanisms, their design has largely prioritised research and data science use cases over real-world clinical workflows. In parallel, practising pharmacists continue to rely on a fragmented landscape of drug information systems, often juggling multiple sources that differ in accessibility, regulatory scope, and clinical depth. This section critically reviews prominent biomedical KGs and pharmacist-facing systems, examining their coverage, update practices, and limitations with respect to local regulatory needs. The discussion highlights a persistent gap between global biomedical knowledge and the practical requirements of pharmacists working within specific national formularies and healthcare systems, which motivates the development of customised, local-aware knowledge graphs.

\subsection{Biomedical Knowledge Graphs in Drug Discovery, Genomic Medicine, and Drug Interactions}\label{sec:bkg-review}

Biomedical Knowledge Graphs (BKGs) integrate heterogeneous biomedical information from public repositories, curated databases, and literature to build richly interconnected resources. These structures enable semantic linkages among drugs, diseases, genes, and phenotypes, powering applications from drug repurposing to precision medicine. Although such graphs incorporate drug-related entities such as indications, interactions, and side effects, they typically prioritise research-driven tasks rather than pharmacist-facing needs.

Several prominent initiatives highlight the evolution of BKG architectures. Early efforts such as Bio2RDF~\citep{belleau2008} demonstrated the feasibility of Linked Data principles for life sciences, integrating genes, proteins, and drugs into a unified RDF framework. However, Bio2RDF has not been updated since 2014, limiting its utility for contemporary pharmacological or clinical decision making.

Hetionet~\citep{himmelstein2017} expanded this paradigm by systematically linking 29 datasets, including DrugBank~\citep{wishart2023}, SIDER~\citep{kuhn2016}, and MEDLINE~\citep{medline2023}, allowing for reasoning about drug-disease, drug-gene and side-effect relationships. Although powerful for discovery applications such as drug repurposing, Hetionet remains static (last released in 2017) and lacks local regulatory information critical for clinical dispensing.

Recent developments have placed greater emphasis on dynamic data ingestion and personalised medicine. SPOKE~\citep{morris2023}, for example, integrates clinical, molecular and wearable device data, covering sources such as LINCS~\citep{koleti2018} and the GWAS Catalog~\citep{sollis2023}. Its continuous update pipeline contrasts sharply with previous static KGs and supports real-time precision health analytics. Similarly, PrimeKG~\citep{chandak2023} aggregates curated resources such as UMLS~\citep{umls2004} and the Mayo Clinic Knowledge Base~\citep{mayoclinic2023}, explicitly modelling indications, contraindications, and phenotypic traits to allow patient-specific predictions.

Specialised graphs have also emerged to address the pharmaceutical and pandemic contexts. DRKG~\citep{ioannidis2020} leverages DrugBank, GNBR, and Hetionet to construct a drug-gene-disease network optimised for embedding-based machine learning, incorporating COVID-19-specific data during the pandemic. KG-COVID-19~\citep{reese2021} represents a related effort, integrating clinical trials, scientific literature, and molecular databases to rapidly create a customised graph for infectious disease research, with iterative updates.

Meanwhile, DisGeNET~\citep{pinero2020} provides a continuously updated repository linking genes to diseases, but with limited direct pharmaceutical coverage, instead focusing on genomic medicine. PharmKG~\citep{zheng2021} similarly targets machine learning in a similar way, integrating drug-gene-disease associations enriched with multi-omic and chemical features, although without clinical safety annotations.

Other initiatives, such as PharmHKG~\citep{asada2021}, pursue hybrid strategies by combining structured data such as, DrugBank, UniProt~\citep{uniprot2023}) with unstructured textual sources such as MeSH~\citep{lipscomb2000} to enhance node expressiveness. This textual integration improves the graph's semantic richness, but remains largely research-oriented rather than oriented toward clinical workflows.

Table~\ref{tab:kg-comparison} summarises the core characteristics of these representative biomedical KGs, comparing their focus areas, data integration strategies, and distinguishing features.

While collectively demonstrating the power of data integration across biomedical domains, none of these graphs is designed for front-line pharmacist use. They generally omit national regulatory authorisation, product-specific labelling, and therapeutic contextualisation necessary for dispensing decisions. Their global or US-centric data focus limits their adaptability to small-state healthcare settings like Malta.

This gap underscores the necessity for pharmacist-centred regulatory-aware knowledge graphs such as \textit{medicX-KG}, which not only integrate drug information but also embed national regulatory data and clinical safety cues. Section~\ref{sec:pharmacist-systems} will further examine how current pharmacist-facing information systems address or fail to address these locality-specific and operational needs.

\begin{table}[h]
\centering
\caption{Comparison of representative Biomedical Knowledge Graphs (BKGs) by scope, primary data sources, update strategy, and distinguishing features.}
\label{tab:kg-comparison}
\begin{tabular}{|p{2.4cm}|p{2cm}|p{2cm}|p{2cm}|p{2.6cm}|}
\hline
\rowcolor{gray!25} \textbf{Knowledge Graph} & \textbf{Focus Area} & \textbf{Primary Data Sources} & \textbf{Update Strategy} & \textbf{Distinguishing Features} \\
\hline
Bio2RDF~\citep{belleau2008} & General biomedicine & Genes, proteins, diseases, drugs & Last updated 2014 & Early pioneer in Linked Data for life sciences; now outdated \\
\hline
\rowcolor{gray!10}DisGeNET~\citep{pinero2020} & Disease genomics & Gene-disease associations from curated and mined sources & Regularly updated & Deep integration in genomics; limited direct pharmaceutical data \\
\hline
Hetionet~\citep{himmelstein2017} & Drug repurposing, disease mechanisms & 29 datasets incl. DrugBank, SIDER, MEDLINE & Static (2017 release) & Broad biomedical entity coverage; strong connectivity modelling \\
\hline
\rowcolor{gray!10}SPOKE~\citep{morris2023} & Precision medicine & LINCS, GWAS, clinical trials, EHRs & Continuously updated & Spans molecular to clinical data; supports real-time analytics \\
\hline
DRKG~\citep{ioannidis2020} & Drug repurposing & DrugBank, GNBR, Hetionet, literature & Static (2020 release) & Embedding-optimised KG; includes COVID-19-specific data \\
\hline
\rowcolor{gray!10}PrimeKG~\citep{chandak2023} & Personalised medicine & UMLS, Mayo Clinic, DrugBank, ontologies & Snapshot (2023) & Emphasizes indications/contraindications and phenotype links \\
\hline
PharmHKG~\citep{asada2021} & Pharmaceutical NLP integration & DrugBank, UniProt, MeSH & Snapshot & Combines structured and textual data for enhanced node richness \\
\hline
\rowcolor{gray!10}PharmKG~\citep{zheng2021} & ML benchmarking, drug-gene-disease inference & DrugBank, gene expression data, literature & Static & Embedding-oriented KG with multi-omic and chemical attributes \\
\hline
KG-COVID-19~\citep{reese2021} & Pandemic response & COVID-19 literature, protein/drug databases, clinical trials & Iteratively updated & Rapid integration; tailored to infectious disease research \\
\hline
\end{tabular}
\end{table}

\subsection{Pharmacist-Facing Systems and Regulatory Gaps}\label{sec:pharmacist-systems}

Although biomedical knowledge graphs (BKGs) have greatly expanded data-driven research in pharmacology and genomics, practising pharmacists rely on operational information systems to support clinical decision-making at the point of care. Key references include the BNF~\citep{bnf2024}, Summary of Product Characteristics (SmPC) and subscription-based clinical support tools such as Micromedex~\citep{micromedex2024} and UpToDate~\citep{uptodate2024}. These resources provide critical guidance on indications, dosages, contraindications, DDIs, and ADRs.

However, such systems operate in silos and often misalign with local regulatory contexts~\citep{fip2023access}. Pharmacists must consult multiple references to compile complete drug profiles, a process confirmed in pilot interviews as time-consuming and error-prone~\citep{flynn2021}. International references such as the BNF list drugs and formulations not authorised locally, while national databases such as the MMA~\citep{mma2024} provide regulatory approval data but limited clinical content.

Fragmentation is further exacerbated by cost barriers. Many pharmacists lack institutional access to premium databases~\citep{nicholson2020}, instead relying on free but inconsistent or outdated resources. The result is a bifurcated information infrastructure, split between global biomedical knowledge and local regulatory registries, with neither fully meeting operational needs.

Partial efforts, such as the \textit{medicX} platform~\citep{farrugia2020}, have addressed aspects of this gap by integrating curated interaction checks into the Maltese context. However, the absence of a fully locality-aware, pharmacist-facing semantic framework embedding national regulatory specifics and therapeutic equivalences remains a critical bottleneck. Closing this gap is essential to deliver context-sensitive, regulation-compliant, and clinically actionable decision support.

\subsection{Limitations in Existing BKGs and Implications for Localised Deployment}\label{sec:limitations}

Despite advances in biomedical knowledge graphs (BKGs), existing resources fall short of pharmacist-facing, locality-sensitive deployments. Several structural deficiencies limit their clinical applicability.

First, most BKGs prioritise broad biomedical discovery without accounting for jurisdictional healthcare differences~\citep{callahan2020}. Resources such as the Drug Repurposed Knowledge Graph (DRKG)~\citep{ioannidis2020} and GNBR~\citep{percha2018} draw heavily from US-centric data, encoding drugs, indications, and guidelines that may not align with other national formularies. In Malta, deploying such graphs risks revealing unauthorised medications or omitting critical local treatments, underscoring the need for jurisdiction-specific tailoring.

Second, sustainability remains a persistent challenge. Many prominent BKGs, such as Bio2RDF~\citep{belleau2008}, are built on static data snapshots and lack active maintenance pipelines, quickly diverging from current pharmacological knowledge. Dynamic resources such as SPOKE~\citep{morris2023} demonstrate continuous updating but require substantial investment~\citep{peng2023}. For pharmacist-facing systems, where clinical safety hinges on real-time accuracy, reliance on outdated graphs is untenable. Sustainable deployments must integrate automated update pipelines from both global databases such as DrugBank~\citep{knox2024} and national authorities.

Third, existing BKGs are conceptually optimised for research discovery rather than operational pharmacy workflows~\citep{hou2023}. They prioritise drug-gene-disease links, but rarely encode practical knowledge such as exhaustive DDIs, pregnancy safety profiles, dosage nuances, or local market availability~\citep{ren2022}. Adapting such graphs for front-line pharmacy use would require substantial re-engineering, including semantic pruning, ontology restructuring, and interface redesign~\citep{callahan2020}.

Thus, there remains a fundamental mismatch between the structural assumptions of current BKGs and the real-world needs of pharmacy practice. Addressing this requires not only adaptation, but the deliberate construction of pharmacist-orientated, locality-aware knowledge graphs designed from first principles.

\subsubsection{Challenges in Biomedical Knowledge Graph Design, Sustainability, and Regulatory Alignment}\label{sec:challenges}

Constructing a pharmacist-oriented biomedical knowledge graph (BKG) capable of locality-aware deployment necessitates overcoming four interdependent challenges: heterogeneous data integration, sustainability, regulatory compliance, and quality assurance.

\textbf{Data Integration and Heterogeneity}: Biomedical knowledge remains fragmented across repositories with varying schemas, identifiers, and terminologies~\citep{fernandez2022}. Integration requires careful entity alignment, schema reconciliation, and semantic consistency maintenance~\citep{himmelstein2017}. For localised deployments, additional complexity arises from reconciling global databases like DrugBank with country-specific regulatory datasets and brand-specific product registries~\citep{zhu2020}.

\textbf{Dynamic Knowledge and Sustainability}: Biomedical data evolves rapidly. New drug approvals, interactions, and safety alerts require continuous updates~\citep{chandak2023}. Without sustained pipelines, even well-constructed graphs lose clinical relevance within months~\citep{bonner2022}. Sustainable BKGs must therefore integrate real-time ingestion pipelines complemented by expert validation workflows~\citep{bikaun2024, lu2025}.

\textbf{Local Regulatory Alignment}: Most existing BKGs omit jurisdiction-specific authorisation status, labelling, or formulary information~\citep{amcp2024}. Localised deployments must explicitly incorporate national constraints, filtering or flagging non-authorised drugs, and encoding region-specific clinical guidelines~\citep{reese2021}. Collaboration with regulatory authorities improves trustworthiness and compliance.

\textbf{Maintaining Quality and Trust}: Clinical applicability requires rigorous quality control. Integration errors, outdated entries, or inconsistent mappings can undermine user trust and endanger patient safety~\citep{scheife2015}. High-quality BKGs require multi-tier validation, expert review of critical entities, and feedback-driven refinement cycles.

Addressing these interconnected challenges is essential to construct sustainable, regulation-compliant, and clinically usable pharmacist-oriented BKGs.

\subsection{Summary and Motivation for medicX-KG}\label{sec:summary-motivation}

Despite their substantial contributions to biomedical discovery, existing BKGs are structurally unsuited for pharmacist-facing, locality-aware clinical applications. Their global orientation, static update strategies, and research-driven priorities often misalign with the operational realities of pharmacy practice~\citep{himmelstein2017, ioannidis2020, morris2023}. 

In fragmented regulatory environments such as Malta, which is characterised by dual alignment with the European Medicines Agency (EMA) and post-Brexit United Kingdom frameworks, language-based pharmaceutical import restrictions, and the absence of a unified national drug repository, general-purpose BKGs fall short to accommodate local drug authorisations, therapeutic substitutions, brand-specific availabilities, and national formulary constraints. As confirmed in previous Section~\ref{sec:pharmacist-systems} and Section~\ref{sec:limitations}, pharmacists are left to manually reconcile global references, which is a fragmented and error-prone process~\citep{flynn2021, nicholson2020}.

Although initiatives such as the original \textit{medicX} platform~\citep{farrugia2020} demonstrated the value of localising drug interaction resources, they did not deliver a fully semantic, dynamic, and regulatory-aligned knowledge infrastructure tailored to national pharmacy needs.

The development of \textit{medicX-KG} is motivated by the need to address these limitations directly. Its conception will therefore be grounded in four strategic priorities:
\begin{itemize}
    \item \textbf{Regulatory Alignment}: Embedding national formulary constraints and approval statuses directly into the KG structure to ensure context-relevant retrieval.
    \item \textbf{Pharmacy-Centric Usability}: Prioritising support for operational queries related to DDIs, therapeutic alternatives, contraindications, and product availability.
    \item \textbf{Localised Predictive Support}: Enabling predictive discovery of DDI within the scope of nationally authorised products, enhancing safety in polypharmacy scenarios.
    \item \textbf{Extensible Semantic Infrastructure}: Architecting a modular schema that accommodates regulatory updates, multilingual expansions, and SmPC parsing over time.
\end{itemize}

Rather than assume global homogeneity, \textit{medicX-KG} intends to embrace the regulatory, linguistic, and clinical diversity inherent in pharmacy practice. It will potentially be an example for next-generation pharmacist-oriented knowledge graphs, which integrate semantically rigorous, dynamically maintainable, and aligned with national regulatory ecosystems.

The technical architecture, ontology design, data integration pipeline and evaluation methodology underpinning \textit{medicX-KG} are presented in the following sections.

\section{Materials and Methods}\label{sec:methodology}

This section outlines the interview study conducted with pharmacists and the construction of the medicX-KG knowledge graph. We detail the qualitative methodology used for interviews (including data analysis), summarise quantitative insights from the interview data, and describe the technical process of building and mapping the knowledge graph with illustrative examples.

\subsection{Qualitative Study: Interviews with Pharmacists}

\subsubsection{Participants, Interview Design, and Procedure} 
To inform the design of medicX-KG with real-world pharmacist needs, we conducted semi-structured interviews with six pharmacists (five female, one male) based in Malta. Participants were purposefully selected to maximise diversity across practice domains, including community pharmacy, hospital paediatrics, regulatory risk management, green manufacturing, quality assurance, and digital health (see Table~\ref{tab:participants}). All interviewees had doctoral degrees in fields related to pharmacy.

\renewcommand{\arraystretch}{1.5}
\begin{table}[h]
\caption{Participant profiles highlighting diverse practice settings. Each pharmacist is assigned a reference code (P1–P6).}\label{tab:participants}
\begin{tabular}{| m{3.5cm} | m{6cm} |>{\centering\arraybackslash}m{2.0cm} |}
\hline
\rowcolor{gray!25} \centering \textbf{Pharmacist's Role} & \centering \textbf{Description} & \textbf{Code} \\
\hline
Green Manufacturing & Sustainable pharmaceutical production expert & P1 \\
\hline
\rowcolor{gray!10} Risk Management Specialist & Pharmaceutical risk assessor and compliance expert & P2 \\
\hline
Community Pharmacy Practice & Medication dispensing and public health advising & P3, P5 \\
\hline
\rowcolor{gray!10} Paediatric/Neonatal Hospital Pharmacy & Specialist in critical care pharmacotherapy for infants and children & P4 \\
\hline
Quality Assurance in Healthcare & Regulator of clinical and pharmaceutical quality systems & P6 \\
\hline
\end{tabular}
\end{table}

\begin{figure}[h]
    \centering
    \includegraphics[width=0.8\textwidth]{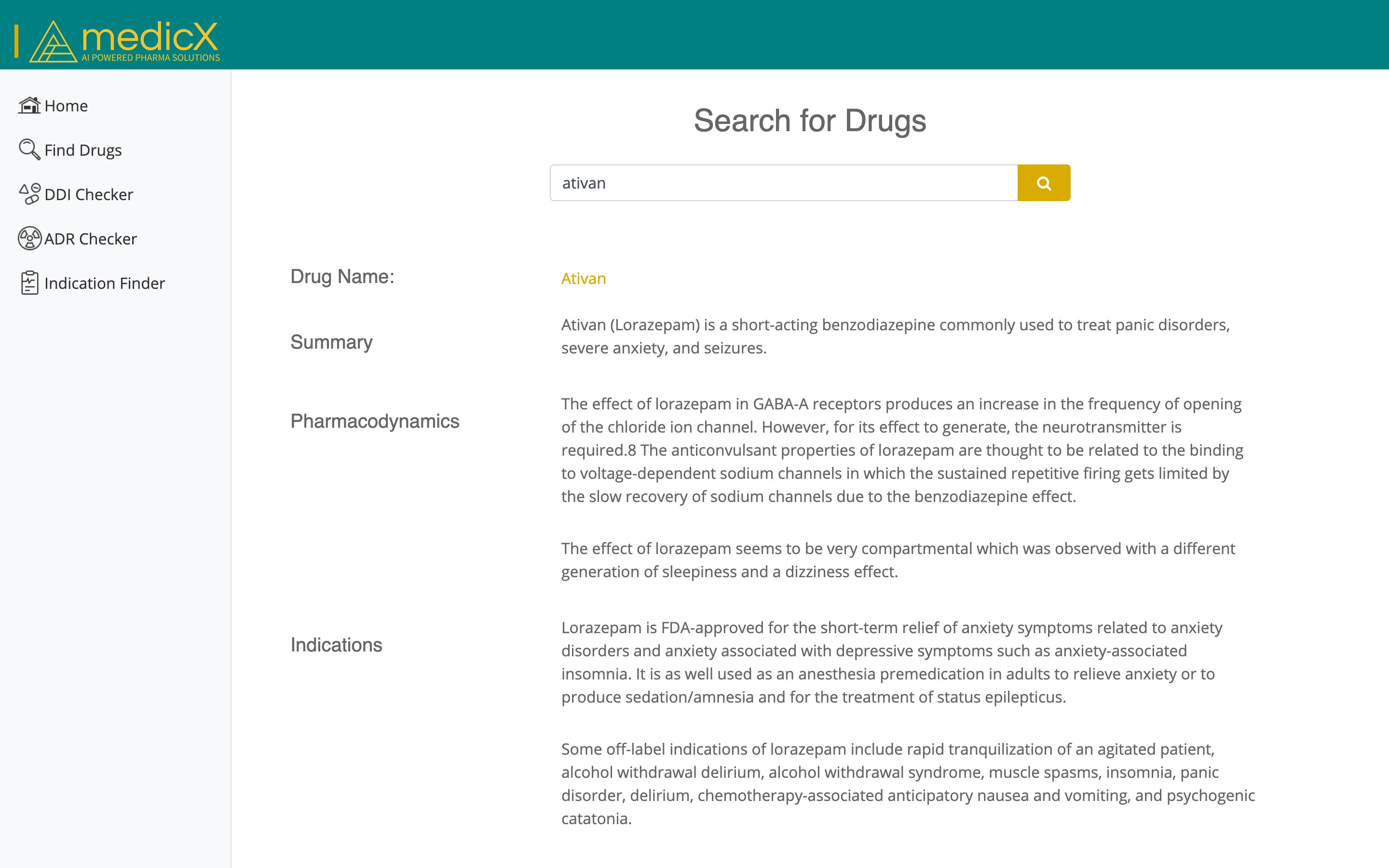}
    \caption{Minimum Viable Product (MVP)~\cite{stevenson2024} of the medicX portal, showing semantic drug search, pharmacodynamics, indications, and predictive DDI scoring.}
    \label{fig:medicx-mvp-drug-info}
\end{figure}

The interviews were conducted remotely (owing to COVID-19 restrictions) by two researchers, including a senior pharmacy expert. Each session lasted approximately 60 minutes, was recorded with consent and complemented by interviewer notes. The interviews followed an open-ended, exploratory format structured around three focal areas:
\begin{enumerate}
    \item \textbf{Information-Seeking Behaviour}: exploration of the types of drug-related information most frequently queried, resources used, and perceived deficiencies.
    \item \textbf{Evaluation of the medicX MVP}: participants interacted with a Minimum Viable Product (MVP)~\citep{stevenson2024} of the medicX portal (Figure~\ref{fig:medicx-mvp-drug-info}), featuring semantic drug search, local SmPC integration, and a predictive DDI checker. The MVP evaluation was framed around usability, relevance, and desired features.
    \item \textbf{Expectations for Predictive DDI Features}: soliciting perspectives on the feasibility, trust, and clinical utility of AI-driven DDI prediction within pharmacy workflows.
\end{enumerate}

The interview guide was developed collaboratively and was informed by previous studies on pharmacist information practices~\citep{romagnoli2016, kontsioti2022, tan2023}. The questions were refined iteratively by consulting with a practising pharmacist to ensure clinical relevance. The final list of questions that were asked during the interview is found in Appendix~\ref{secA1}.

\subsubsection{Qualitative Data Analysis Methodology} 

All interviews were transcribed verbatim. We used an inductive thematic analysis approach~\citep{braun2006}, allowing dominant patterns to emerge directly from the data without forcing predefined categories.

Two researchers independently performed open coding\footnote{In qualitative thematic analysis, coding refers to the analytic process of identifying, labelling, and organising features of the data that are relevant to answering the research question. Codes capture meaningful characteristics of the data and form the foundation for later theme development}, annotating key statements and observations across the dataset. Coding outcomes were then iteratively reviewed and clustered into higher-order thematic abstractions through collaborative discussion and consensus-building.

Coding was organised around three initial anticipated dimensions, namely drug information needs, information resources and improvement suggestions; however, thematic refinement led to expanded categories such as ``local drug availability challenges'' and ``trust in predictive features.'' Frequency counts between participants were used to quantify theme relevance where applicable.

The qualitative coding process was enhanced through the following:
\begin{itemize}
    \item \textbf{Domain-expert triangulation}: involving a pharmacist researcher in all stages of the analysis to validate domain-specific interpretations.
    \item \textbf{Collaborative codebook refinement}: codes were merged, split, or redefined across multiple review cycles to maintain semantic precision.
    \item \textbf{Cross-participant comparison}: examining consistency or divergence of themes in different settings of pharmacy practice.
\end{itemize}

This analytic process enabled the identification of critical pharmacist needs, including gaps in existing drug information systems, high trust thresholds for predictive features, and strong demand for local regulatory alignment, which directly informed the subsequent knowledge graph schema and query mechanisms developed for medicX-KG.

\subsection{Findings from Interviews: Key Information Needs and Resource Use}\label{sec:interview_findings}

Thematic analysis of the interviews revealed critical pharmacist information needs, resource usage patterns, and priority areas for system improvement (see Appendix~\ref{secA3} for more information about the findings from the interviews). Although the sample size (N=6) was limited, the diversity in practice settings enhances the indicative value of the findings. However, the results should be interpreted as indicative rather than representative. 

\subsubsection{Frequently Sought Drug Information Types}
\begin{figure}[h]
    \centering
    \includegraphics[width=\textwidth]{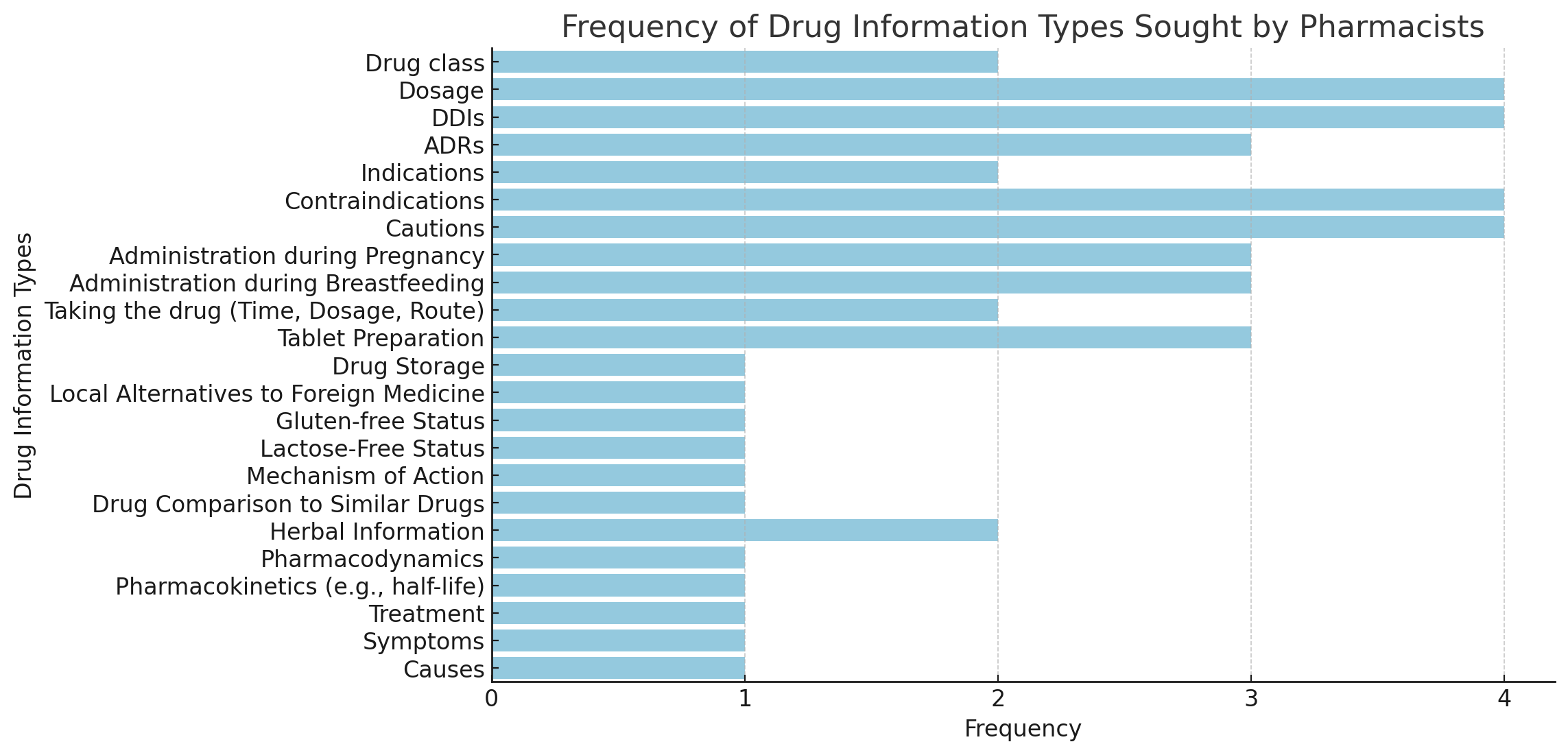}
    \caption{Frequently sought drug information types (number of mentions across six participants).}
    \label{fig:info-types-bar}
\end{figure}

Pharmacists consistently prioritised information related to dosage, DDIs, contraindications, and cautions, each mentioned by four of the six participants (Figure~\ref{fig:info-types-bar}). ADRs were also prominent, highlighted by half of the interviewees. 

Lower frequency but clinically significant queries included drug alternatives, pharmacokinetics, dietary compatibility (e.g. gluten-free medications) and pregnancy/lactation safety. Such specialised information needs often arise in context-specific problem-solving scenarios.

\begin{figure}[ht]
    \centering
    \includegraphics[width=0.7\textwidth]{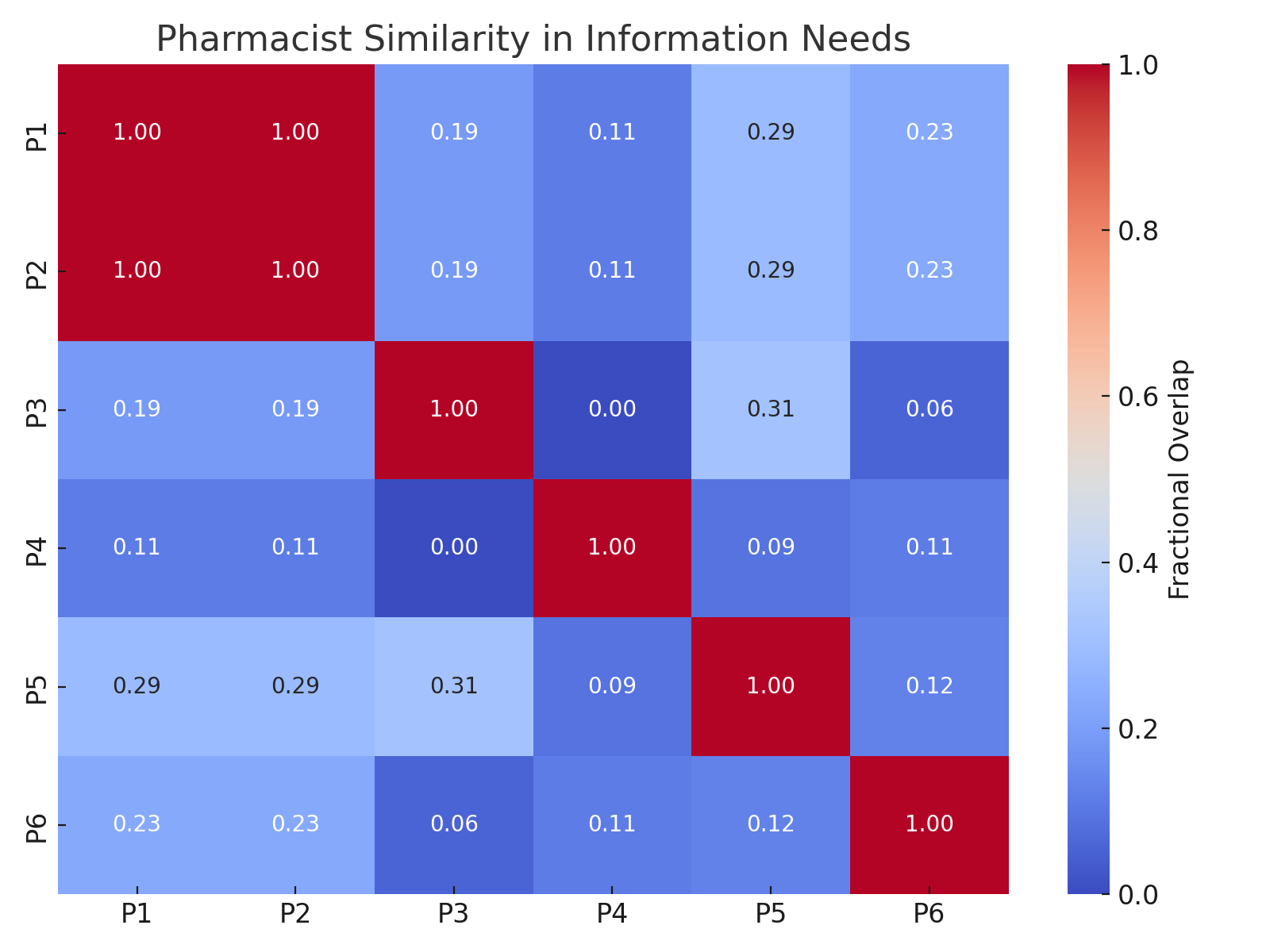}
    \caption{Heatmap of information need similarity between participants (Jaccard index).}
    \label{fig:heatmap-similarity}
\end{figure}

Pairwise similarity analysis (Figure~\ref{fig:heatmap-similarity}) further revealed convergence among pharmacists in similar roles (e.g., community pharmacy), and divergence among specialists, underscoring the heterogeneous nature of information needs across practice contexts.

\subsubsection{Drug Information Resource Usage and Gaps}
SmPC documents and the BNF~\citep{bnf2024} were the most consistently used references, cited by all the pharmacists. International commercial databases such as Micromedex~\citep{micromedex2024} and UpToDate~\citep{uptodate2024} were also widely consulted (5/6), although cost barriers limited their accessibility.

\begin{figure}[ht]
    \centering
    \includegraphics[width=0.9\textwidth]{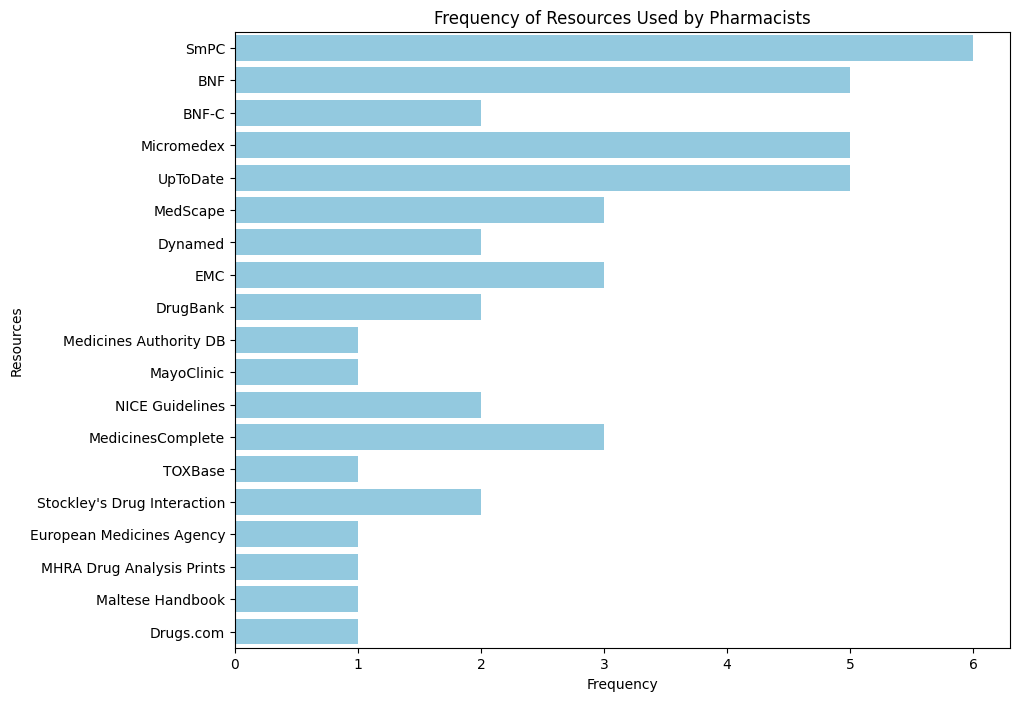}
    \caption{Primary drug information resources cited by pharmacists.}
    \label{fig:drug-resource-bar}
\end{figure}

Other resources such as MedScape~\citep{medscape2024}, EMC~\citep{emc2024}, DynaMed~\citep{dynamed2024}, and the MMA’s online platform were mentioned sporadically. In particular, several pharmacists reported resorting to ad-hoc online searches when authoritative resources were inaccessible, raising concerns about the reliability of the information.

The fragmented and non-standardised ecosystem of references, ranging from highly credible SmPCs to ad hoc Google searches, underscores the need for a unified, authoritative knowledge graph consolidating trusted sources.

\subsubsection{Feedback on medicX MVP and Desired Features}
Pharmacists expressed strong support for core aspects of the medicX portal (more information related to the MVP is found in Appendix~\ref{secA4}) but also identified clear areas for improvement:
\begin{itemize}
    \item \textbf{Information Presentation}: Calls for ADRs to be presented in frequency-ranked tables with clear definitions (e.g., 'common' side effects explicitly quantified).
    \item \textbf{Usability Enhancements}: Suggestions for interactive navigation features (e.g., clickable sections, collapsable menus) and mobile-friendly layout adaptations.
    \item \textbf{Terminology Simplification}: Emphasis on avoiding jargon and clarifying clinical terms to ensure quick and safe comprehension.
    \item \textbf{Clinical Decision Support Tools}: Recommendations for integrated calculators (e.g., dosage adjustment by indication) and drug–food/herb interaction alerts.
    \item \textbf{Local Regulatory Visibility}: Requests for explicit labelling of local authorisation status (prescription vs. over-the-counter, brand equivalence).
\end{itemize}

The feedback notably converged on the need for:
\begin{itemize}
    \item Trustworthy, concise, contextually relevant information.
    \item Rapid accessibility through optimised UI/UX design.
    \item Local regulatory contextualisation embedded within drug profiles.
\end{itemize}

These findings directly informed medicX-KG’s schema, data requirements (e.g., ADR frequency attributes, local regulatory metadata), and interface design priorities.

\subsection{Knowledge Graph Construction and Ontology Mapping}

Following the needs assessment from the interviews, we constructed the medicX-KG, which is tailored to address the identified information needs and integrate the disparate sources of drug information. This section provides technical details on the data sources used, the ontology design and mapping strategy to merge these sources, and examples of how mapping ambiguities were resolved to ensure a consistent and reliable graph.

\subsubsection{Data Sources}\label{sec:data_sources}
medicX-KG integrates three primary datasets selected to balance regulatory, clinical, and molecular information needs: the MMA registry~\citep{mma2024}, the BNF\citep{bnf2024}, and DrugBank\citep{knox2024}. While earlier sections reviewed the characteristics of these sources, here we focus on the technical extraction procedures and design choices underpinning their integration.

The MMA dataset provided the national reference list of pharmaceutical products authorised in Malta. The programmatic harvest of the online MMA portal yielded 9,784 product records. Pre-processing eliminated structurally invalid entries, withdrawn products, and duplicates, producing a curated subset of 9,746 products suitable for downstream integration. Each retained record contained structured fields specifying the product name, active ingredients (with concentrations), pharmaceutical form, and marketing authorisation holder.

BNF data, accessed via an academic licence, offered structured clinical information for more than 1,600 drug entries. Extraction focused on fields deemed most critical for pharmacist decision-making, including indications, standard dosage regimens, contraindications, ADRs, and DDIs. Web scraping was implemented using Selenium and BeautifulSoup to extract these fields from the online BNF repository, preserving both textual content and structured subsections.

To supplement and enrich the KG with molecular and classification data, DrugBank version 5.1.4 was incorporated. This resource provided chemical structures, pharmacodynamics, ATC codes, and extensive synonym dictionaries for 14,315 drugs. Extraction of DrugBank entries prioritised fields essential for cross-dataset entity resolution and pharmacological classification. Unlike the MMA and BNF sources, DrugBank was partially pre-structured; however, minor inconsistencies (e.g., legacy synonyms, deprecated ATC codes) were addressed during harmonisation.

All datasets underwent rigorous cleaning and standardisation prior to integration. The names of the products and ingredients were normalised for casing, punctuation, and unit representation. Synonym fields were expanded through controlled vocabularies and ambiguous cases were flagged for manual curation during mapping. Where multiple synonyms or formulations were encountered for a single compound, primary names were selected based on regulatory precedence or international non-proprietary nomenclature (INN) standards. More information related to the data used from each of the sources can be found in the Appendix~\ref{secA2}.

This phase was not purely preparatory but a critical enabler of the semantic interoperability and clinical fidelity that medicX-KG aimed to achieve. The design decisions taken during data curation, including preserving granular distinctions between salt forms and formulations while unifying pharmacological equivalences, would later shape the ontology structure and mapping logic.

\subsubsection{Mapping Strategy}~\label{sec:mapping_strategy}
The central technical challenge in building medicX-KG was the accurate mapping of each product and active ingredient from the MMA dataset to the corresponding entries in the BNF and DrugBank.

Although established ontology alignment systems such as LogMap~\citep{logmap2011}, OLaLa~\citep{hertling2023}, and Matcha-DL~\citep{cotovio2024} offer state-of-the-art capabilities for schema matching and logical alignment, their applicability to pharmaceutical datasets proved limited. While recent advances such as OLaLa have improved matching performance in settings with sparse logical structure by leveraging large language models, and Matcha-DL has enhanced scalability and multilingual support, these tools remain primarily optimised for reconciling structured OWL ontologies where class hierarchies and formal axioms are consistently available. In contrast, pharmaceutical product datasets, such as those from regulatory sources, are characterised by semi-structured formats, noisy terminologies, incomplete semantics, and domain-specific idiosyncrasies~\citep{shvaiko2013,hertling2022}.

Consequently, we adopted a custom rule-based mapping strategy deliberately designed to maximise interpretability, modularity, and domain-specific control, which are essential criteria in pharmaceutical contexts, where regulatory compliance, clinical accuracy, and manual traceability are paramount. This approach aligns with precedents in the construction of biomedical knowledge graphs, such as Hetionet~\citep{himmelstein2017} and DRKG~\citep{ioannidis2020}, which similarly eschewed generic matchers in favour of domain-aware and heuristic-based reconciliation pipelines.

Our mapping procedure was implemented as a multistage, rule-based process comprising four core stages:

\begin{enumerate} \item \textbf{Direct Name Matching:} For each active ingredient in the MMA dataset, we first attempted an exact match against BNF drug names, applying normalisation for case, punctuation, and common unit descriptors. Products such as \textit{Zyrtec 10mg tablets} (active ingredient: Cetirizine Hydrochloride) were successfully linked via this straightforward approach.

\item \textbf{Synonym and Salt Resolution:} Where direct matches failed, we leveraged DrugBank’s extensive synonym and salt-form mappings. For example, the MMA listing for \textit{Esomeprazole Magnesium} was matched to the BNF monograph for \textit{Esomeprazole} after identifying the salt relationship via DrugBank synonym data.

\item \textbf{Combination Product Decomposition:} For combination drugs (e.g., \textit{Amoxicillin + Clavulanic acid}), we implemented a decomposition logic. Where there was a combined BNF entry (e.g., \textit{Co-amoxiclav}) existed, mapping was performed accordingly. Otherwise, products were individually assigned to their component monographs, ensuring retrieval fidelity across different representation styles.

\item \textbf{Unique Identifier Assignment:} To maintain source traceability and prevent collisions, each \texttt{Product} entity in the graph was assigned a URI derived from standardised product names, pharmaceutical forms and, where available, marketing authorisation codes.

\end{enumerate}

\begin{table}[h]
\centering
\caption{BNF Mapping Outcomes for MMA Products and Components. This table summarises the results of mapping active ingredient names and product combinations from the MMA dataset to BNF drug entries. The matching was performed in stages: direct string match, component decomposition for combination drugs, and synonym/salt resolution via DrugBank. Failures are also reported when no equivalent BNF entry could be identified even after normalisation or synonym expansion.}
\label{tab:bnf-mapping}
\begin{tabular}{|p{8cm} | p{4cm}|}
\hline
\rowcolor{gray!25}\textbf{Mapping Outcome} & \textbf{Number of Products or Components} \\
\hline
Direct match to BNF entry (e.g., "Cetirizine hydrochloride") & 677 \\
\hline
\rowcolor{gray!10}No match to BNF (even after normalisation) & 852\\
\hline
Mapped via component decomposition (BNF entries exist for parts of combination) & 114 \\
\hline
\rowcolor{gray!10}Component-level mapping failed (no BNF entries found for any components) & 180 \\
\hline
Mapped via synonym/salt (from DrugBank) to BNF & 17  \\
\hline
\rowcolor{gray!10}Synonym-based mapping failed (BNF entry still missing) & 331\\
\hline
\end{tabular}
\end{table}

Mapping outcomes are summarised in Tables~\ref{tab:bnf-mapping}–\ref{tab:pubchem-mapping}. In the first phase, BNF matching yielded 677 direct alignments, supplemented by 114 matches through decomposition strategies. However, 852 entries remained unmatched after exhaustive resolution of synonyms and salt forms against BNF alone, illustrating the structural gaps in cross-national drug information coverage.

\begin{table}[h]
\centering
\caption{DrugBank Mapping Outcomes for Active Ingredients and Components. This table summarises the outcome of attempts to match MMA-derived drug ingredients to DrugBank entries using direct string matching, synonym resolution, and salt normalisation. It also includes mappings at component level for combination products not covered in BNF, and rare fallback mappings using full product names when ingredient-level data was unavailable.}
\label{tab:drugbank-mapping}
\begin{tabular}{|p{8cm} | p{4cm}|}
\hline
\rowcolor{gray!25}\textbf{Mapping Outcome} & \textbf{Number of Products or Components} \\
\hline
Mapped via direct match, synonym, or salt normalisation & 1,061  \\
\hline
\rowcolor{gray!10}No match to DrugBank (after synonym/salt attempt) & 468\\
\hline
Component-level mapping to DrugBank (for combinations not found in BNF) & 194 \\
\hline
\rowcolor{gray!10}Mapped via full product name only (ingredient not found in any DB) & 11  \\
\hline
\end{tabular}
\end{table}

The integration of DrugBank proved pivotal, resolving 1,061 additional mappings, particularly for international brand variants, novel entities, and alternative formulations absent from BNF. Combination mappings and fallback matching based on complete product names further improved coverage. The remaining unmatched products, most often veterinary medications or highly localised formulations, were addressed through PubChem, achieving 1,226 chemical level matches, although with 303 entities still unresolved due to the lack of canonical identifiers.

\begin{table}[h]
\centering
\caption{PubChem Mapping Outcomes Used as a Chemical Identifier Fallback. This table presents the results of mapping remaining unmatched drug entities (after BNF and DrugBank attempts) to PubChem compounds. PubChem served as a final tier resource for resolving chemical identifiers when all other strategies failed.}
\label{tab:pubchem-mapping}
\begin{tabular}{|p{8cm} | p{4cm}|}
\hline
\rowcolor{gray!25}\textbf{Mapping Outcome} & \textbf{Number of Products or Components} \\
\hline
Direct match to PubChem compound & 1,226 \\
\hline
\rowcolor{gray!10}No match in PubChem (exhausted all mapping tiers) & 303 \\
\hline
\end{tabular}
\end{table}

\begin{figure}[h]
    \centering
    \includegraphics[width=0.9\linewidth]{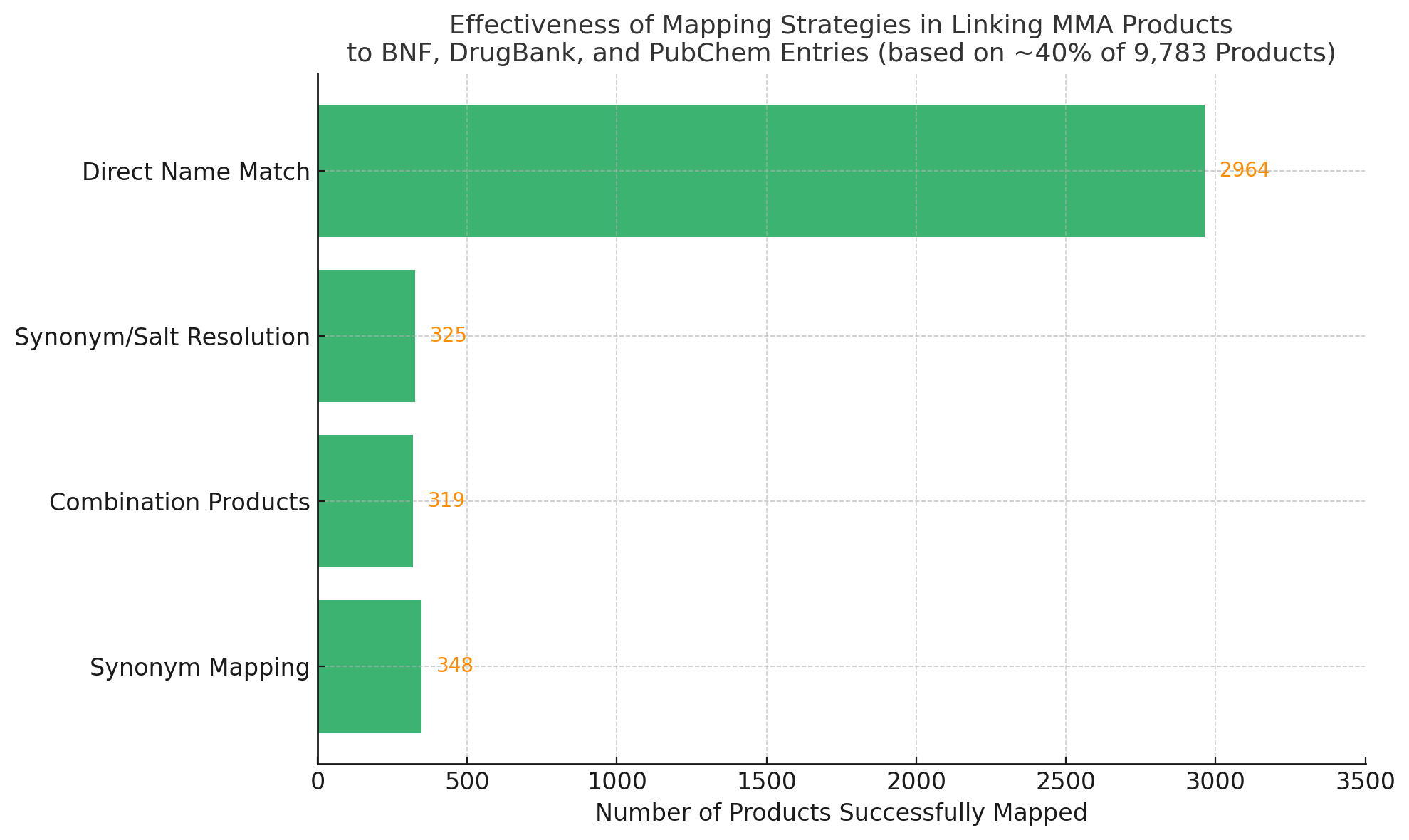}
    \caption{Effectiveness of Mapping Strategies in Linking MMA Products to BNF and DrugBank Entries. The bars correspond to distinct mapping strategies applied during entity resolution, not individual data sources. Counts reflect successfully mapped products using each technique, sometimes involving multiple sources (e.g. DrugBank used for salt resolution followed by BNF linking).}
    \label{fig:mapping-strategies}
\end{figure}

The layered effectiveness of the mapping pipeline is further illustrated in Figure~\ref{fig:mapping-strategies}, which aggregates the mappings by conceptual reconciliation strategy rather than by source. This abstraction highlights that integration success was not source-dependent, but method-dependent, emerging from cumulative, cross-source techniques operating in tandem.

Concrete examples illustrate the criticality of the multistage pipeline. For example, the MMA listing \textit{Zyrtec 10mg tablets} was easily mapped by direct matching to the BNF entry for \textit{Cetirizine Hydrochloride}. In contrast, \textit{Augmentin 500mg/125mg tablets} required decomposition into its active components, \textit{Amoxicillin} and \textit{Clavulanic acid}, followed by synonym-based linking via DrugBank to the BNF monograph for \textit{Co-amoxiclav}. More complex cases such as \textit{Adrenaline} (British English) vs. \textit{Epinephrine} (international naming) required a synonym expansion informed by DrugBank and PubChem to achieve the correct alignment of the entity.

Thus, the mapping strategy in medicX-KG was not a preparatory step but a core scientific contribution, enabling accurate integration of regulatory, clinical, and chemical knowledge. It established the semantic foundations necessary for the ontology design detailed in Section~\ref{sec:ontology_design}, and for the subsequent clinical usability of the knowledge graph in real-world pharmacist settings.

To ensure persistence, disambiguation, and traceability across reconciled entities, each product in medicX-KG was subsequently assigned a globally unique URI, constructed from standardised product attributes (e.g. name, pharmaceutical form, marketing authorisation codes). In parallel, the mapping strategy preserved formulation specificity at the \texttt{Product} level, capturing differences between capsules, tablets, injectables, while unifying shared pharmacological properties at the \texttt{Active\_Ingredient} layer. This layered abstraction was crucial for supporting fine-grained clinical queries without redundancy and laid the structural foundation for the ontology design.

\subsubsection{Ontology Design}\label{sec:ontology_design}

The construction of medicX-KG involved a tightly integrated process of systematic data transformation and ontology schema design, both guided by the need to support pharmacist decision-making in regulated environments. Particular emphasis was placed on ensuring semantic precision, clinical relevance, regulatory traceability, and extensibility for future biomedical expansions.

\paragraph{Data Transformation Process}

Following entity mapping and reconciliation (Section~\ref{sec:mapping_strategy}), curated pharmaceutical datasets were transformed into a formal semantic representation using the Resource Description Framework (RDF). A custom-built \textit{RDFizer} tool, leveraging the Python RDFLib library,\footnote{\url{https://rdflib.dev/}} was developed to automate this process. The tool parsed structured records from the Medicines Authority portal, BNF, and DrugBank, applied normalization procedures (e.g., unit standardization, casing harmonization, synonym disambiguation), and serialized the resulting data into RDF triples.

The transformation pipeline was designed to promote semantic interoperability and compliance with the FAIR (Findable, Accessible, Interoperable, Reusable) data principles~\citep{wilkinson2016}. Provenance was explicitly maintained through attribution properties, enabling every assertion within the KG to be traceable to its original source. Although the underlying architecture is complete, public release is pending the development of a continuous learning mechanism and further validation within pharmacist workflows, as outlined in Section~\ref{sec:future_work}.

\paragraph{Ontology Schema Design}

\begin{figure}[h]
    \centering
    \includegraphics[width=0.85\textwidth]{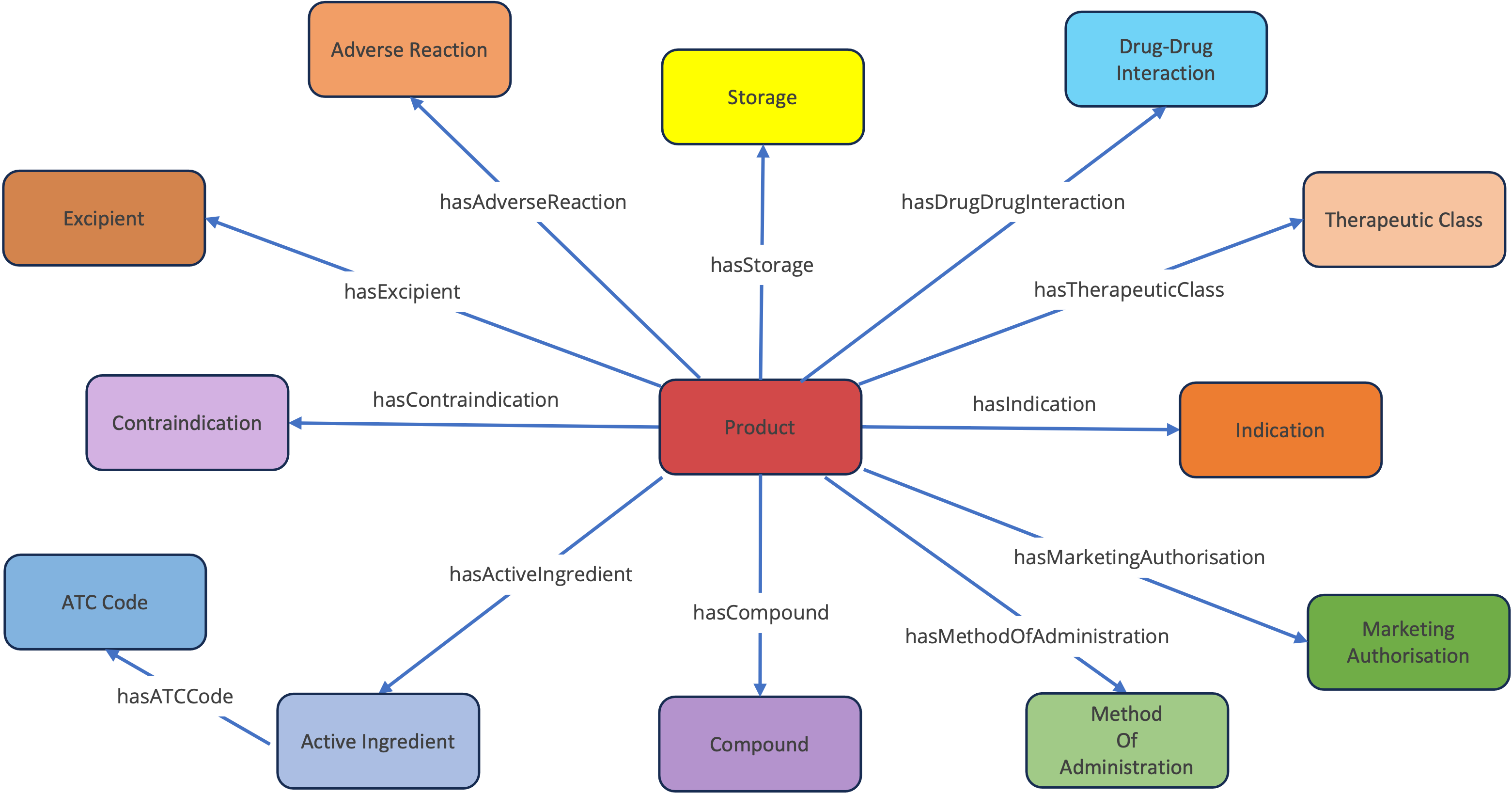}
    \caption{High-level T-Box schema of medicX-KG, illustrating the principal classes and object properties that underpin the semantic structure of the knowledge graph. The ontology captures clinical, pharmacological, and regulatory information essential for pharmacist decision support in regulated environments.}
    \label{fig:kg-ontology}
\end{figure}

The design of the medicX-KG ontology was structured to satisfy three principal objectives: (i) capturing clinical and pharmacological expressiveness relevant to pharmacist workflows, (ii) supporting semantic interoperability and property-rich reasoning, and (iii) enabling modular extensibility. A high-level representation of the medicX-KG T-Box schema, capturing its principal classes and object properties, is provided in Figure~\ref{fig:kg-ontology}.

At the core of the ontology, distinct classes represent:

\begin{itemize}
    \item \texttt{Product}: Specific medicinal products authorised in Malta, distinguished by formulation and strength.
    \item \texttt{ActiveIngredient}: Chemical substances (INNs) encapsulating shared pharmacological properties.
    \item \texttt{Compound}: Additional chemical constituents present within formulations.
    \item \texttt{Excipient}: Non-active ingredients included in drug formulations.
    \item \texttt{Indication}: Therapeutic conditions targeted by medicinal products.
    \item \texttt{Contraindication}: Clinical scenarios where product use is contraindicated.
    \item \texttt{AdverseDrugReaction}: Documented adverse events linked to product usage.
    \item \texttt{TherapeuticClass}: Pharmacological groupings supporting therapeutic substitution.
    \item \texttt{ATCCode}: Anatomical Therapeutic Chemical classification codes.
    \item \texttt{Storage}: Regulatory and clinical storage requirements.
    \item \texttt{MarketingAuthorisation}: Authorisation entities granting market approval.
    \item \texttt{MethodOfAdministration}: Approved methods for administering drugs.
    \item \texttt{DrugDrugInteraction}: Entities capturing pharmacokinetic or pharmacodynamic interactions between two or more active ingredients.
\end{itemize}

Each \texttt{Product} instance links semantically to its associated \texttt{ActiveIngredient}, as well as to clinical and regulatory concepts such as \texttt{Indication}, \texttt{Contraindication}, \texttt{AdverseDrugReaction}, \texttt{Storage} requirements, and \texttt{MethodOfAdministration}. \texttt{ActiveIngredient} nodes are cross-referenced with \texttt{ATCCode} classifications, supporting therapeutic grouping and substitution. Furthermore, \texttt{DrugDrugInteraction} nodes explicitly model clinically significant interactions, annotated with interaction mechanisms, clinical effects, and severity levels.

Unlike simpler pharmacist-facing datasets that capture drug attributes individually, medicX-KG models complex pharmacological and regulatory realities through semantically rich n-ary relationships. This enables more expressive queries, such as identifying contraindicated product combinations or retrieving storage-specific regulatory conditions, thus supporting evidence-based decision making in both clinical and regulatory contexts.

Critically, the ontology design of medicX-KG was deliberately aligned with the dual practical challenges pharmacists encounter: ensuring therapeutic appropriateness while adhering to regulatory frameworks. By integrating clinical entities (e.g., indications, ADRs, contraindications) with regulatory artefacts (e.g., marketing authorisations, storage requirements), medicX-KG provides a unified semantic infrastructure tailored to pharmacist needs. This design is particularly pertinent for smaller jurisdictions such as Malta, where pharmacists must navigate partially harmonised EU frameworks alongside non-EU supply chains, such as UK imports following Brexit. The ontology remains operationally lightweight for real-world usability while enabling structured, regulation-aware knowledge exploration.

\paragraph{Design Principles}

Several key design choices underpin the semantic robustness and extensibility of medicX-KG:

\begin{itemize}
    \item \textbf{Formulation granularity:} \texttt{Product} nodes preserve dosage form specificity (e.g., capsule, suspension, injection), critical for dispensing accuracy.
    \item \textbf{Salt normalization:} Variants are normalized at the \texttt{ActiveIngredient} level, with clinically relevant salt descriptors preserved at the \texttt{Product} level.
    \item \textbf{Therapeutic class linkage:} \texttt{ActiveIngredient} nodes are cross-linked to \texttt{ATCCode} entries, facilitating therapeutic substitution workflows.
    \item \textbf{Interaction modelling:} Drug-drug interactions are modelled as first-class entities, enabling property-rich annotation including severity, mechanism, and evidence source.
    \item \textbf{Evidence provenance:} All assertions retain explicit source attributions (e.g., MMA portal, BNF, DrugBank), ensuring transparency and auditability.
    \item \textbf{Modular extensibility:} The ontology is designed to accommodate future domains such as pharmacogenomics and real-world evidence without structural disruption.
\end{itemize}

In summary, the medicX-KG ontology and data transformation framework provide a structured, clinically relevant, and regulation-aware representation of pharmaceutical knowledge tailored to pharmacist decision support. The following sections present the evaluation results and discuss the clinical applicability, strengths, and limitations of the knowledge graph.

\section{Results and Discussion}\label{sec:results}
This section presents the evaluation results of the medicX-KG constructed, focusing on its ability to meet pharmacist-facing information needs and its comparative advantages over existing biomedical knowledge graphs. We report both qualitative and quantitative results: first, through an analysis of Competency Questions (CQs) designed to test the practical utility of the KG, and second, through entity and relationship statistics that illustrate the scope and structure of the graph. Finally, we discuss how medicX-KG compares with broader knowledge graphs highlighting its unique value in terms of jurisdictional specificity, clinical relevance, and pharmacist-oriented design.

\subsection{Competency Questions Evaluation}\label{sec:cq-evaluation}
Task–oriented validation of ontologies and knowledge graphs usually follows the CQ pattern, whereby domain experts first craft natural‑language questions that capture the intended scope, translate them into SPARQL queries, and then check whether the graph returns complete and correct answers. We have adopted this methodology (analysed and formalised by \citet{keet2016}) to evaluate the effectiveness of \textit{medicX-KG}.

We defined a set of CQs that mirror common pharmacist information needs. These questions span categories such as drug dosage, product availability, DDIs, ADRs, therapeutic indications, and safety in special populations. Table~\ref{tab:cq-eval} summarises seven representative CQs, their thematic focus, and the extent to which medicX-KG currently answers them. The SPARQL representation of the CQs as well as other information related to the evaluation can be found in Appendix~\ref{secA5}.

The evaluation involved executing each CQ on a test set of drug entities within the KG and verifying the accuracy and completeness of the retrieved answers against established references (e.g. SmPCs, BNF).

\begin{table}[h]\centering
\caption{Summary of competency queries (CQs) evaluated on \textit{medicX-KG}. \textit{Fully Met} indicates that the KG provides a complete and accurate answer; \textit{Partially Met} denotes partial or incomplete answers; \textit{Not Met} signifies inability to answer under current KG content.}
\label{tab:cq-eval}
\begin{tabular}{|p{4cm} | p{1.5cm} | p{1.5cm} | p{4cm}|}
\hline
\rowcolor{gray!25}\textbf{Competency Question (CQ)} & \textbf{Category} & \textbf{Outcome} & \textbf{Notes} \\
\hline
CQ1: What is the recommended dosage of \textit{drug X}? & Dosage & \textit{Not Met} & Detailed posology absent; SmPC dosage data not integrated. \\
\hline
\rowcolor{gray!10}CQ2: Which authorised products contain \textit{drug X} in Malta? & Availability & \textit{Fully Met} & Complete listing based on MMA regulatory data. \\
\hline
CQ3: Are there known interactions between \textit{drug X} and \textit{drug Y}? & DDI & \textit{Fully Met} & Retrieved with severity levels from BNF and DrugBank. \\
\hline
\rowcolor{gray!10}CQ4: Which adverse drug reactions are associated with \textit{drug X}? & ADR & \textit{Fully Met} & Standard ADR profiles captured. \\
\hline
CQ5: For which conditions is \textit{drug X} indicated? & Indication & \textit{Partially Met} & Broad indications available; disease-specific mapping incomplete. \\
\hline
\rowcolor{gray!10}CQ6: Which other drugs share the same therapeutic class as \textit{drug X}? & Therapeutic Class & \textit{Fully Met} & Resolved via ATC code-based classification. \\
\hline
CQ7: Can \textit{drug X} be used during pregnancy or breastfeeding? & Precaution & \textit{Partially Met} & Basic safety flags available; detailed SmPC precautions missing. \\
\hline
\end{tabular}
\end{table}

The results demonstrate that medicX-KG is already capable of answering a substantial subset of pharmacist-relevant queries. Availability (CQ2), DDI (CQ3), ADR (CQ4), and therapeutic class queries (CQ6) are consistently \textit{fully met}, showcasing the strengths of the KG’s structured integration of MMA product lists, BNF clinical content, and DrugBank pharmacological data. 

For example, CQ2 queries returned comprehensive results: the KG correctly retrieved all 52 amoxicillin-containing products authorised in Malta, including branded formulations like \textit{Augmentin} and \textit{Moxilen}. Similarly, DDI evaluations (CQ3) demonstrated not only the identification of known interactions, such as amoxicillin’s effect on warfarin anticoagulation, but also the retrieval of clinically significant recommendations (e.g., INR monitoring).

ADR retrievals (CQ4) were also robust. Querying common ADRs for amoxicillin yielded a complete list aligned with standard references, including hypersensitivity reactions, gastrointestinal disturbances, and dermatological effects.

However, questions demanding finer-grained or unstructured clinical knowledge exposed gaps. CQ1, regarding dosage guidelines, was \textit{not met}, reflecting the current absence of structured posology information within the KG. Likewise, CQ5 and CQ7 were only \textit{partially met}: while major therapeutic indications and safety flags are modelled, detailed condition-specific guidance and comprehensive pregnancy/breastfeeding precautions are lacking.

Tables~\ref{tab:entity-stats} and~\ref{tab:rel-stats} contextualise the CQ evaluation outcomes by quantifying the size and scope of the knowledge graph. The high number of instances of DDIs (1.39M) and ADR (2,703) explains the robustness observed in CQ3 and CQ4, while the absence of dose-specific entities highlights why CQ1 remains unmet.

\begin{table}[h]\centering
\caption{Entity statistics in \textit{medicX-KG}, reflecting the scale and composition of integrated drug, clinical, and regulatory information.}
\label{tab:entity-stats}
\begin{tabular}{|l | r| }
\hline
\rowcolor{gray!25}\textbf{Entity Class} & \textbf{Instances} \\
\hline
Active Ingredient & 5,085 \\
\hline
\rowcolor{gray!10}ATC Code & 5,313 \\
\hline
ADR & 2,703 \\
\hline
\rowcolor{gray!10}Contraindication & 1,459 \\
\hline
Drug-Drug Interaction & 1,390,406 \\
\hline
\rowcolor{gray!10}Indication & 2,787 \\
\hline
Product & 9,747 \\
\hline
\rowcolor{gray!10}Therapeutic Class & 256 \\
\hline
\end{tabular}
\end{table}

\begin{table}[h]\centering
\caption{Relation statistics in \textit{medicX-KG}, representing pharmacological, clinical, and regulatory linkages between entities.}
\label{tab:rel-stats}
\begin{tabular}{|l | r| }
\hline
\rowcolor{gray!25}\textbf{Relation Type} & \textbf{Instances} \\
\hline
\texttt{has\_active\_ingredient} & 12,577 \\
\hline
\rowcolor{gray!10}\texttt{has\_active\_ingredient\_dosage} & 12,577 \\
\hline
\texttt{has\_atc} & 4,858 \\
\hline
\rowcolor{gray!10}\texttt{has\_contraindication} & 3,105 \\
\hline
\texttt{has\_drug\_interaction} & 2,780,806 \\
\hline
\rowcolor{gray!10}\texttt{has\_indication} & 3,973 \\
\hline
\texttt{has\_side\_effect} & 384,490 \\
\hline
\rowcolor{gray!10}\texttt{has\_therapeutic\_class} & 1,118 \\
\hline
\end{tabular}
\end{table}

Importantly, the statistics reflect the full entity and relationship inventory, including data points indirectly derived from external sources such as PubChem and DrugBank synonym expansions, not solely MMA-authorised products. This breadth partially explains the robust performance of KG for ADR and interaction queries.

The partial fulfilment of the therapeutic indication (CQ5) and precaution (CQ7) questions stems from the KG’s reliance on ATC categories and high-level safety flags, without exhaustive incorporation of detailed clinical narratives typical of SmPC. Similarly, absence of granular dosage recommendations prevents the KG from addressing CQ1. These limitations are consistent with the strategic prioritisation during KG construction (Section~\ref{sec:mapping_strategy}), which emphasised relational modelling over free text extraction.

Taken together, the results demonstrate that medicX-KG already achieves strong competency in structured relational knowledge retrieval critical to pharmacist workflows, particularly product identification, interaction management and ADR recognition, while highlighting key opportunities for future enrichment in dosage guidance, detailed indications, and special population safety profiling.

\subsection{Comparing medicX-KG with Biomedical Knowledge Graphs and Clinical Databases}\label{sec:kg-comparison}
A robust evaluation of \textit{medicX-KG} requires its positioning relative to established biomedical knowledge graphs (KGs) and clinical databases. Table~\ref{tab:medicx-kg-comparison} summarises key comparative features against four prominent resources: DrugBank~\citep{wishart2023}, Hetionet~\citep{himmelstein2017}, PharmKG~\citep{zheng2021}, and Micromedex~\citep{micromedex2024}. The comparison is structured around regulatory specificity, clinical usability, and functional complementarity, which together define medicX-KG's distinct contribution.

\begin{table}[!htbp]
\centering
\caption{Comparative Analysis of Biomedical Knowledge Graphs and Clinical Databases Relevant to medicX-KG.}
\label{tab:medicx-kg-comparison}
\begin{tabular}{|p{2cm}|p{1.8cm}|p{1.6cm}|p{1.6cm}|p{1.7cm}|p{1.8cm}|}
\hline
\rowcolor{gray!25} 
\textbf{Aspect} & \textbf{DrugBank} & \textbf{Hetionet} & \textbf{PharmKG} & \textbf{Micromedex} & \textbf{medicX-KG} \\
\hline
Primary Purpose & Pharmacolog-ical reference & Translational research (drug repurposing) & Biomedical relation mining & Clinical decision support  & Pharmacist-oriented KG for local markets \\
\hline
\rowcolor{gray!10}Regulatory Info & No & No & No & Partial (FDA focus)  & Yes (Malta-specific MMA integration) \\
\hline
Jurisdictional Modelling & No & No & No & No  & Yes (national authorisations and product forms) \\
\hline
\rowcolor{gray!10}Clinical Usability & Limited & No & No & Yes  & Yes (structured pharmacist-facing entities) \\
\hline
ADR Structuring & Partial (textual) & No & Partial (relation mining) & Yes (structured)  & Yes (structured, frequency categorised) \\
\hline
\rowcolor{gray!10}DDI Structuring & Partial & Partial & Partial & Yes (structured)  & Yes (structured, local product-aware) \\
\hline
Open Access & Yes (academic license) & Yes & Yes  & No (commercial license) & Planned (post-validation) \\
\hline
\rowcolor{gray!10}Biomedical Enrichment & Yes (ATC, pharmacodynamics) & Yes (genes, pathways) & Yes (biomedical triplets) & Limited & Yes (via DrugBank, ATC codes) \\
\hline
Graph-Based Queryability & Limited (flat database) & Yes & Yes & No  & Yes (fully RDF-based KG) \\
\hline
\rowcolor{gray!10}Designed for Pharmacists & No & No & No & Partial & Yes (clinically actionable pharmacist use cases) \\
\hline
\end{tabular}
\end{table}

\subsubsection{Regulatory Specificity and Local Market Integration}

As Table~\ref{tab:medicx-kg-comparison} highlights, \textit{medicX-KG} uniquely integrates jurisdiction-specific regulatory metadata, a dimension entirely absent from DrugBank, Hetionet, and PharmKG. Unlike these KGs, which abstract drug knowledge without reference to national healthcare systems, medicX-KG anchors pharmaceutical entities to the Malta Medicines Authority (MMA)~\citep{mma2024} authorisation records. This allows direct querying of locally available formulations, a critical requirement for real-world pharmacist decision-making.

Micromedex partially addresses regulatory approval via FDA-focused content but lacks subnational granularity and local market adaptation. By contrast, medicX-KG operationalises regulatory specificity at a national scale, bridging the critical gap between theoretical pharmacology and practical dispensability.

\subsubsection{Alignment with Pharmacist-Facing Clinical Needs}

Clinical decision support demands not just biomedical coverage but structured, context-sensitive knowledge. General-purpose KGs like Hetionet and PharmKG prioritise research-driven relationships (e.g., drug–gene–disease triplets), leaving point-of-care questions underserved. DrugBank provides pharmacokinetics and adverse effect descriptions but offers limited structured decision support linked to national products.

Micromedex addresses pharmacist workflows more directly but, as Table~\ref{tab:medicx-kg-comparison} shows, its knowledge model is proprietary, non-graph-based, and lacks open semantic interoperability.

In contrast, \textit{medicX-KG} systematically encodes pharmacist-critical entities,  including pregnancy/breastfeeding cautions, renal/hepatic dose adjustments, frequency-categorised ADRs, allergy warnings, and DDIs, all linked explicitly to jurisdiction-approved products. The KG architecture supports complex queries enabling clinical insights unattainable through generic KGs.

\subsubsection{Positioning and Complementarity}

\textit{medicX-KG} is not designed to replace general biomedical KGs but to complement them. It extends molecular and pharmacological standards imported from DrugBank and ATC taxonomies while layering a regulatory and pharmacist-centric semantic model missing from traditional resources.

Compared to Micromedex, medicX-KG offers three distinct advantages: open (planned) access, native graph-based querying, and national regulatory integration. This positions it as a novel hybrid KG, simultaneously aligned with international biomedical standards and adapted to jurisdictional healthcare realities.

The competency evaluation (Section~\ref{sec:cq-evaluation}) empirically validates this complementarity: medicX-KG answered pharmacist-specific queries around availability, contraindications, and ADR risks with a precision and semantic richness unmatched by broader KGs or commercial databases.

In summary, \textit{medicX-KG} demonstrates a new paradigm: embedding regulatory fidelity and clinical expressiveness into an interoperable graph framework optimised for pharmacist use at the point of care. Its architecture is readily extensible to other small-market jurisdictions, offering a scalable model for bridging global biomedical resources with local clinical needs.

\section{Conclusion and Future Work}\label{sec:conclusion}

The development of \textit{medicX-KG} represents a significant advance toward a pharmacist-centric knowledge representation for drug information. By harmonising data from the MMA, DrugBank, and the BNF, medicX-KG consolidates both international pharmacological knowledge and jurisdiction-specific regulatory details into a unified semantic infrastructure. This integration addresses persistent gaps in pharmacists' access to actionable drug information, particularly in small-state healthcare systems such as Malta.

A key contribution of medicX-KG lies in its regulatory alignment and task-specific design. Unlike broader biomedical knowledge graphs that focus on research discovery, medicX-KG is explicitly oriented toward front-line pharmacy practice. The graph supports queries around drug availability, therapeutic alternatives, DDIs, ADRs, and clinical precautions. Its structure was iteratively informed by pharmacist interviews and competency question evaluations, resulting in a resource that demonstrably improves information accessibility for core pharmacy tasks.

The evaluation confirmed that medicX-KG effectively answers pharmacist-relevant queries, particularly in domains where structured relationships, such as DDIs, ADRs, and regulatory authorisations, are available. By consolidating fragmented sources into a pharmacist-facing, semantically enriched system, medicX-KG demonstrates the feasibility and necessity of locality-aware knowledge graphs in clinical support roles.

\subsection{Limitations}

Although medicX-KG marks an important step, several limitations warrant acknowledgement. First, the current graph does not encode detailed dosage schedules or administration guidelines, as SmPC posology data have not yet been systematically integrated. Consequently, the KG is less effective in answering precise dosing or titration queries.

Second, medicX-KG is built on a static data snapshot. Although authoritative at the time of extraction, the KG lacks a real-time update mechanism, risking information staleness as drug authorisations change or new interaction knowledge emerges. 

Third, KG has not yet been subjected to longitudinal field testing within operational pharmacy settings. Although interviews and MVP evaluations offer promising indications, real-world deployment is essential to validate usability, interface design, integration with pharmacy management systems, and impact of clinical decisions.

These limitations highlight the need for further iteration in both content enrichment and operational integration.

\subsection{Future Work}\label{sec:future_work}

In our future research on \textit{medicX-KG} we intend to proceed along three tightly coupled strands: data dynamism, predictive intelligence, and validation at the point of care. Each designed to consolidate the KGs scientific reliability and clinical utility.

To prevent semantic drift, we plan to develop a continuous‑integration (CI) pipeline that incrementally ingests MMA data, DrugBank releases, and BNF updates. Building on KG streaming maintenance techniques that interleave entity resolution with graph-delta application~\citep{Rossanez2020,Xu2020}, the pipeline will (i) fingerprint incoming records, (ii) resolve identity and provenance conflicts through a supervised matcher, and (iii) run regression tests that rerun the full competency question (CQ) suite after every data refresh.  

The enriched, versioned KG will serve as substrate for machine‑explainable drug–drug‑interaction (DDI) inference. We will extend our earlier explainable link‑prediction framework~\citep{farrugia2023} by encoding DDIs as RDF* reifications with mechanism, strength, and evidence attributes, thereby allowing graph-based transformers over typed paths to output both a risk score and a mechanistic narrative trace. Aligning with best practice in clinical AI~\citep{guidotti2019}, every predicted interaction will be accompanied by provenance links to metabolic routes, shared targets, or co‑reported adverse events.

We intend to perform improve CQ validation process using a large‑scale, publicly released benchmark. Drawing on the PrimeKGQA~\citep{yan2024} template library, we intend to generate dozens of parametrised CQs that span all major entity types. Every answer will be annotated independently by at least two pharmacists and adjudicated, with agreement reported via Cohen’s~$\kappa$ and Krippendorff’s~$\alpha$. As evaluation metrics we will consider ranked‑retrieval measures (precision, recall, $F_{1}$), thereby capturing partial correctness and answer ordering. It is also possible to automatically compile Natural‑language CQs into SPARQL‑OWL using the translator of ~\cite{fernandez2022}, ensuring syntactic fidelity and lowering maintenance overhead.

Usability work will converge on a pharmacist‑centric front‑end that combines SPARQL‑backed widgets with retrieval‑augmented generation (Graph‑RAG) to permit free‑text querying without sacrificing traceability. Pilot installations in community and hospital settings will measure task‑completion time, error rate, and System Usability Scale scores, following the study protocol of \cite{hyzy2022}. Feedback loops from these pilots will feed directly into ontology extensions, based on multilingual SmPC literals in Maltese, and into adaptive ranking models that privilege locally authorised products.

Finally, we will generalise the ontology modules that encode jurisdictional constraints so that regulators in small EU markets such as Cyprus, Ireland, can supply a mapping file and instantiate their own national view of the KG. For cross‑jurisdiction alignment we can reuse the SHACL‑based constraint system proposed by \cite{robaldo2024} to guarantee conformance with EMA identifiers while allowing local formulary overrides.

In summary, by coupling automated data stewardship with explainable inference and statistically robust, user‑centred evaluation, the next iteration of medicX-KG aims to become a continuously learning, regulator‑aligned knowledge infrastructure that demonstrably improves the safety and efficiency of medication management.

\clearpage

\backmatter

\begin{appendices}

\section{Questionnaire}\label{secA1}

\subsection{Participant Information Seeking Methods}
\begin{itemize}
    \item What specific types of drug-related information do you frequently need to search for in your professional practice?\\
    \item When you need information related to pharmacodynamics, which sources do you typically consult? Please list them.\\
    \item How successful are you in finding the information you need through these sources?\\
    \item On a scale of 1 to 5 (1 being extremely difficult and 5 being very easy), how would you rate the ease of finding the necessary information through your chosen sources?\\
    \item In your experience, what are some gaps or limitations you've encountered in existing platforms or information sources when searching for medicinal information?
\end{itemize}

\subsection{Seeking feedback related to the medicX-prototype}
\begin{itemize}
    \item What features of the medicX portal did you find particularly useful or appealing?\\
    \item Were there any features of the medicX portal that you didn't find helpful or disliked?\\
    \item Are there any features you would recommend removing or modifying in the medicX portal? Please provide details on how you think they could be improved.\\
    \item Did you feel that the level of detail provided in medicX was sufficient for your information needs? If not, what additional information would you like to see?\\
    \item Are there any other features or functionalities you believe should be added to the medicX portal to enhance its utility for healthcare professionals like yourself?
\end{itemize}

\subsection{DDI Predictive Feature}
\begin{itemize}
    \item Regarding the Drug-Drug Interaction (DDI) predictive feature in medicX, would you trust its predictions? Please elaborate on your level of trust.\\
    \item What factors or information would be necessary for you to have full trust in the DDI predictive feature within medicX?
\end{itemize}
\pagebreak
\section{Collected Information}\label{secA2}
\subsection{Malta Medicine's Authority}\label{mma-dataset}
Below is the information that was available in the MMA dataset.
\begin{itemize}
    \item Medicine Name
    \item Active Ingredients (incl. dosage)
    \item Pharmaceutical Forms
    \item Therapeutic Class
    \item Classification: otc, pom
    \item ATC Code
    \item Status
    \item Authorisation Number
    \item Authorisation Date
    \item Authorisation Holder
    \item Authorisation Holder Address
\end{itemize}
\subsection{BNF data that was considered}\label{bnf-dataset}
\begin{itemize}
    \item Medicine Name
    \item Indications
    \item Side effects
    \item Drug-Drug interactions
    \item Pregnancy related data
    \item Breastfeeding guidelines
    \item Allergies
    \item Contraindications
    \item Therapeutic classification
    \item Important safety information
    \item Drug action
    \item Caution
    \item Hepatic impairment
    \item Renal impairment
    \item Patient carer advice
\end{itemize}
\subsection{DrugBank data that was considered}\label{db-dataset}
\begin{itemize}
    \item Medicine description
    \item ATC codes
    \item Targets
    \item Drug-Drug interactions
    \item Enzymes
    \item Transporters
    \item Carriers
    \item Narrow therapeutic index
    \item Food interactions
\end{itemize}
\clearpage

\section{Interview results: Drug-related Information and Drug Resources}\label{secA3}
\begin{table}[h]
\caption{Drug-related information that each of the six interviewees seeks regularly.} \label{tabc1}
\begin{tabular}{| p{0.4\linewidth} | p{0.05\linewidth} | p{0.05\linewidth} | p{0.05\linewidth} |p{0.05\linewidth} | p{0.05\linewidth} | p{0.05\linewidth} |}
\hline
\textbf{Drug Information}  & \textbf{P1} & \textbf{P2} & \textbf{P3}   & \textbf{P4}   & \textbf{P5}   & \textbf{P6} \\ \hline
Drug Class  &   &   & X    &    & X  &  \\ \hline
Dosage   & X   & X   &    & X  & X &     \\ \hline
Drug-Drug Interactions  & X  & X   & X   &  & X &    \\ \hline
Adverse Drug Reactions  &  &  & X   &  & X    &  X  \\ \hline
Indications   &  &   &   & X   &   &  X  \\ \hline
Contraindications   & X  & X   & X &   & X &   \\ \hline
Cautions   & X  & X   & X   &  &  X  &   \\ \hline
Administration during Pregnancy   & X  & X   &   &    &   & X   \\ \hline
Administration during Breastfeeding  & X  & X  &   &    &  & X \\ \hline
Taking the Drug (Time, Dosage, Route)  &   &    &  &  & X  & X \\ \hline
Tablet Preparation  & X   & X   &   &   &    & X   \\ \hline
Drug Storage   &    &    &  &   & X &   \\ \hline
Local Alternatives for Foreign Medicines  &  &  &  X &   &    &  \\ \hline
Gluten-Free Status   &  &  & X &  &   &  \\ \hline
Lactose-Free Status  &  &  & X &  &  &   \\ \hline
Mechanism of Action  &    &   &    &   & X  &  \\ \hline
Drug Comparison to Similar Drugs  &  &   &    &    &  X &  \\ \hline
Herbal Information  & X & X &  &   &   &   \\ \hline
Pharmacodynamics   &    &  &   &  &   & X  \\ \hline
Pharmacokinetics  &   &  &    &     &  & X  \\ \hline
Treatment  &  &  &  X &    &    &  \\ \hline
Symptoms   &  &  &  X  &  &   &  \\ \hline
Causes  &  &  & X &    &   &   \\ \hline
\end{tabular}
\end{table}

\begin{table}[h]
\caption{Drug Resources used by each of the six interviewees on a regular basis.} \label{tabc3}
\begin{tabular}{|p{0.3\linewidth}| p{0.05\linewidth}| p{0.05\linewidth}| p{0.05\linewidth}| p{0.05\linewidth}| p{0.05\linewidth}| p{0.05\linewidth}|}
\hline
\textbf{Resources}                       & \textbf{P1} & \textbf{P2} & \textbf{P3}   & \textbf{P4}   & \textbf{P5}   & \textbf{P6}   \\ \hline
SmPC                                     & X               &X           & X             & X            & X          & X\\ \hline
BNF                                      & X               &X           & X             & X             & X         &                          \\ \hline
BNF-C                                    &                 &            & X             & X             &           &                          \\ \hline
Micromedex                               & X               & X          & X             & X             &           & X                         \\ \hline
UpToDate                                 & X               &            & X             & X             &           & X                         \\ \hline
MedScape                                 &                 &            & X             & X             & X         &                          \\ \hline
Dynamed                                  & X               & X          &               &               &           &                          \\ \hline
EMC                                      & X                & X         & X             &               &           &                          \\ \hline
DrugBank (for laboratory-based research) & X                & X          &              &               &          &                          \\ \hline
Medicines Authority DB                   &                  &           & X             &                &          &                          \\ \hline
MayoClinic                               &                 &            & X             &                &          &                          \\ \hline
NICE Guidelines                          &                  &           & X             & X              &          &                          \\ \hline
MedicinesComplete                        &                  &           & X             & X             &           & X                     \\ \hline
TOXBase                                  &                  &           & X             &               &           &                          \\ \hline
Stockley's Drug Interaction              &                  &           &               & X             &           & X     \\ \hline
European Medicines Agency                &                  &           &               & X             &           &                          \\ \hline
MHRA Drug Analysis Prints                &                  &           &               & X             &           &                          \\ \hline
Maltese Handbook                         &                 &            &               &               & X         &                          \\ \hline
Drugs.com                                &                 &            &               &               &           & X \\ \hline
\end{tabular}
\end{table}

\pagebreak
\begin{longtable}{p{0.3\linewidth} p{0.6\linewidth}}
\caption{During the interviews the group of Pharmacists provided information related to several resources that they frequently consult during their work.}\label{tab4}\\
\hline
\textbf{Resource} & \textbf{Description} \\ \hline
Summary of Product Characteristics (SmPC) & All interviewees indicated that they can access most SmPCs. SmPCs are essential documents that provide detailed information about the properties and usage of medicinal products. Their widespread availability ensures that healthcare professionals have access to comprehensive drug information. \\ \hline
BNF  & All interviewees mentioned that they use the BNF. The BNF is a trusted reference   source for healthcare professionals, providing information on drug indications, dosages, and interactions.  \\ \hline
BNF for Children (BNF-C) & Interviewees confirmed that they use the  BNF-C, which is specifically tailored to paediatric medicine, ensuring that healthcare professionals have accurate and age-appropriate drug information for children. \\ \hline
Micromedex & Most   interviewees reported using Micromedex. Micromedex is a reliable drug reference tool that offers a wide range of drug information, including dosages, interactions, and safety details. \\ \hline
UpToDate  & Interviewees also use UpToDate, a clinical decision support tool that provides evidence-based medical information. It covers various medical topics, including drug-related content.  \\ \hline
MedScape  & Several interviewees mentioned using MedScape, a medical resource platform that offers drug information, medical news, and clinical tools. \\ \hline
Dynamed & One interviewee indicated using DynaMed, a clinical reference tool that provides evidence-based information, including drug-related content. \\ \hline
Electronic Medicines Compendium (EMC) & EMC is another resource that some interviewees utilise. It offers SmPCs and patient information leaflets for medicines authorised in the UK. \\ \hline
DrugBank  & One interviewee mentioned using DrugBank, a database that includes detailed drug information for laboratory and research purposes. \\ \hline
Medicines Authority Website  & Interviewees rely on the Medicines Authority Website, likely for regulatory and authorisation-related information on medicines in Malta. \\ \hline
Mayo Clinic  & Mayo Clinic, a reputable medical resource, was mentioned by one interviewee as a source for drug-related information. \\ \hline
NICE Guidelines (National Institute for Health and Care Excellence) & NICE guidelines were utilised by some interviewees. These guidelines provide evidence-based recommendations for healthcare professionals, including drug prescribing and management. \\ \hline
MedicinesComplete  & MedicinesComplete, which includes the Martindale Drug Reference, BNF and other resources, is used by interviewees for comprehensive drug information. \\ \hline
TOXBase & One interviewee relies on TOXBase, likely for toxicology-related information on drugs and substances. \\ \hline
Stockley's Drug Interaction & Interviewees mentioned using Stockley's Drug Interaction, a reference source specifically focused on drug interactions. \\ \hline
European Medicines Agency (EMA) & EMA was mentioned as a source for regulatory and authorisation information related to   medicines in the European Union. \\ \hline
MHRA Drug Analysis Prints  & Interviewees use MHRA Drug Analysis Prints, which may contain information on the analysis of drugs, including quality and safety assessments. \\ \hline
Maltese Handbook & One interviewee referred to a Maltese Handbook as a resource, possibly for localised drug information and guidelines. \\ \hline
Drugs.com & Interviewees use Drugs.com, an online resource providing drug information for healthcare professionals and consumers. \\ \hline
\end{longtable}

\clearpage

\section{medicX: Minimum Viable Product}\label{secA4}
The concept of an MVP, as described by \cite{stevenson2024}, emphasises rapid learning through a Build-Measure-Learn feedback loop with minimal effort and is central to lean startup methodology, which focuses on quickly testing value propositions to understand customer needs in specific markets \citep{dennehy2019,cook2023}.
 
The medicX MVP was developed using a Python/Django backend and a GraphDB\footnote{\url{https://graphdb.ontotext.com/}} triplestore for semantic data storage. The frontend was implemented using React. Although the system is not yet publicly deployed, development is ongoing with the goal of releasing a beta version. The backend architecture adheres to key FAIR principles—including data provenance and persistent, resolvable identifiers. An open-source release is currently under consideration. In the meantime, the core components of \textit{medicX-KG}, including the structured data and mapping functions, are being made available via GitHub.

The beta version of the \textit{medicX} portal will eventually be hosted at \url{https://medicini.ai}, where \emph{medicini} is the Maltese term for medicines.

\begin{figure}[h!]
\centering
\includegraphics[width=0.78\textwidth]{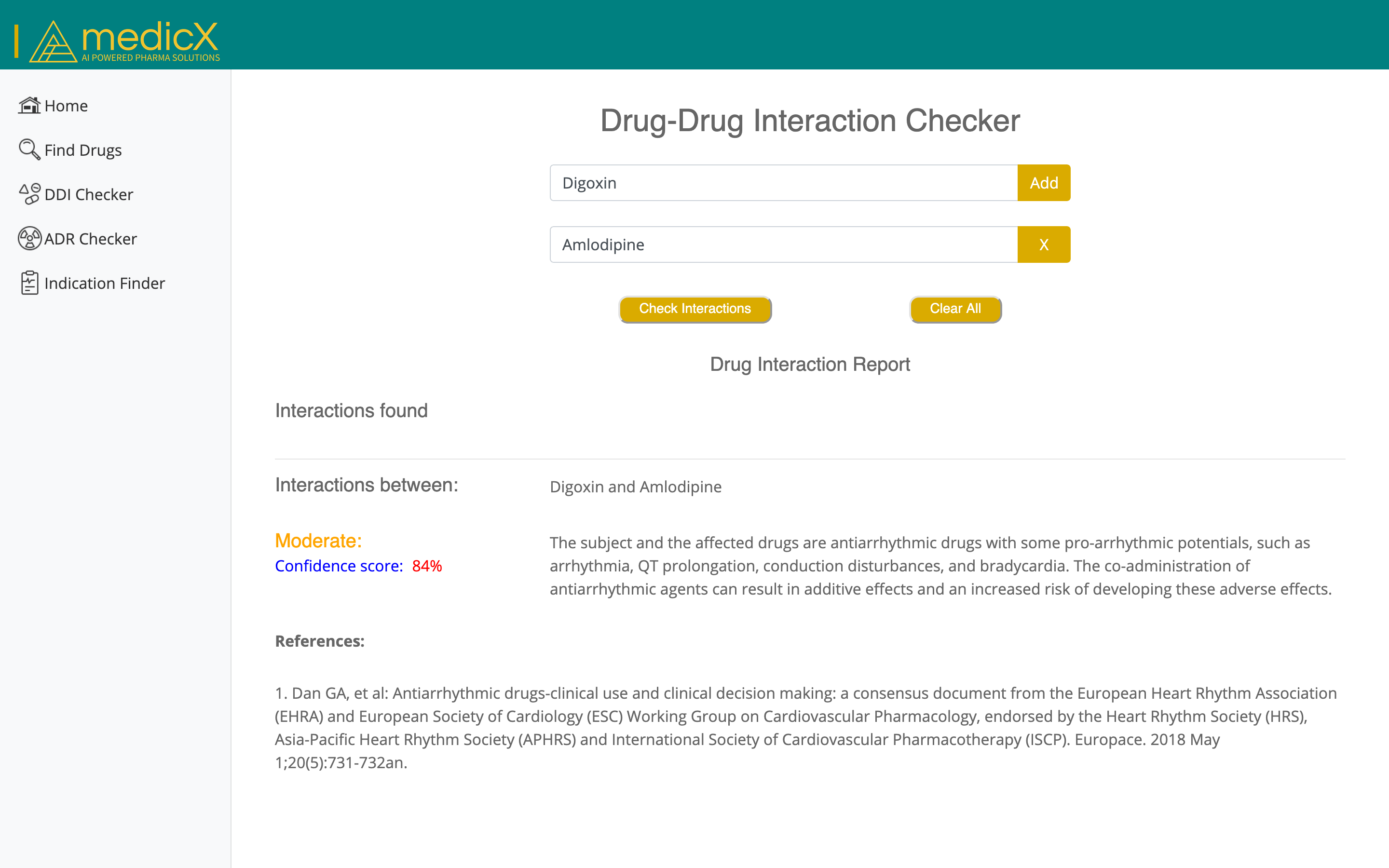}
\caption{The medicX prototype includes facility for DDIs. The information related to interactions between two or more drugs can be checked. In this example, Digoxin and Amlodipine are checked. The result shows that a moderate interaction has been found. A confidence score is computed based on the information available within the medicX-KG as well as references to research that has been extracted from PubMed.}\label{fig:medicx-mvp-ddi}
\end{figure}

\begin{figure}[h!]
\centering
\includegraphics[width=0.78\textwidth]{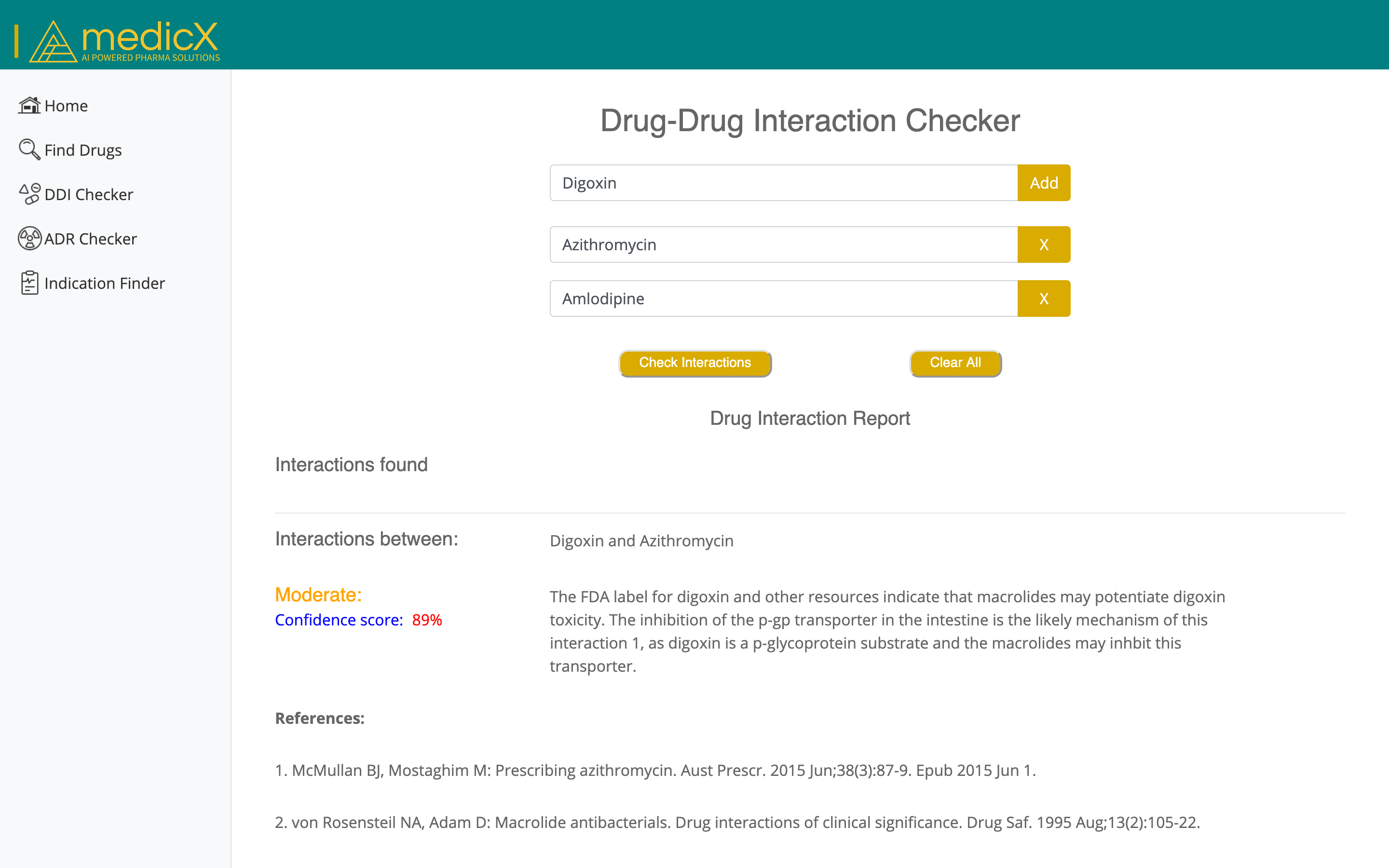}
\caption{In this example, interactions between three drugs (Digoxin, Azithromycin and Amlodipine) are checked. In this case, the result includes information and confidence scores related to the interactions between pairs of drugs (Digoxin and Azithromycin, Digoxin and Amlodipine and, Azithromycin and Amlodipine).}\label{fig:medicx-mvp-multiple}
\end{figure}

\clearpage

\section{Competency Questions}\label{secA5}

\begin{enumerate}

    \item[CQ1.] \textbf{What is the recommended dosage of \textit{$<$drug x$>$}?}
    \begin{lstlisting}[language=SPARQL, caption={SPARQL query for retrieving recommended dosage for a drug}]
        PREFIX mdx: <http://medicX.org/>

        SELECT DISTINCT ?formulation ?dosageValue ?dosageUnit
        WHERE {
            mdx:drugX mdx:has_formulation ?formulation.
            ?formulation mdx:has_dosage ?dosage.
            ?dosage mdx:value ?dosageValue ;
                    mdx:unit ?dosageUnit.
        }
        ORDER BY ?formulation
    \end{lstlisting}

    \item[CQ2.] \textbf{Which authorised products contain \textit{$<$active\_ingredient x$>$} in Malta?}
    \begin{lstlisting}[language=SPARQL, caption={SPARQL query for retrieving authorised products with a specific active ingredient}]
        PREFIX mdx: <http://medicX.org/>
        
        SELECT DISTINCT ?productName ?authorisationStatus
        WHERE {
            ?product mdx:has_active_ingredient mdx:ingredientX.
            ?product mdx:name ?productName.
            ?product mdx:authorisationStatus ?authorisationStatus.
            FILTER (?authorisationStatus = "Authorized")
        }
        ORDER BY ?productName
    \end{lstlisting}

    \item[CQ3.] \textbf{Are there known interactions between \textit{$<$drug x$>$} and \textit{[*{drug}]}?}
    \begin{lstlisting}[language=SPARQL, caption={SPARQL query for interactions between drug x and a set of drugs}]
        PREFIX mdx: <http://medicX.org/>
        
        SELECT DISTINCT ?drugInSet ?interactionType ?interactionSeverity
        WHERE {
            VALUES ?drugInSet { mdx:drugY mdx:drugZ mdx:drugS mdx:drugT }
            mdx:drugX mdx:has_drug_interaction ?interaction.
            ?drugInSet mdx:has_drug_interaction ?interaction.
            ?interaction mdx:interactionType ?interactionType.
            OPTIONAL { ?interaction mdx:interactionSeverity ?interactionSeverity. }
        }
        ORDER BY ?interactionSeverity ?drugInSet
    \end{lstlisting}

    \item[CQ4.] \textbf{Which adverse drug reactions (side effects) are associated with \textit{$<$drug x$>$}?}
    \begin{lstlisting}[language=SPARQL, caption={SPARQL query for retrieving side effects of a drug}]
        PREFIX mdx: <http://medicX.org/>
    
        SELECT DISTINCT ?sideEffectName ?sideEffectSeverity
        WHERE {
            mdx:drugX mdx:has_side_effect ?sideEffect.
            ?sideEffect mdx:name ?sideEffectName.
            OPTIONAL { ?sideEffect mdx:severity ?sideEffectSeverity. }
        }
        ORDER BY ?sideEffectSeverity ?sideEffectName
    \end{lstlisting}

    \item[CQ5.] \textbf{For which conditions or diseases is \textit{$<$drug x$>$} indicated?}
    \begin{lstlisting}[language=SPARQL, caption={SPARQL query for retrieving indications for a drug}]
        PREFIX mdx: <http://medicX.org/>

        SELECT DISTINCT ?indicationName
        WHERE {
            mdx:drugX mdx:has_indication ?indication.
            ?indication mdx:name ?indicationName.
        }
        ORDER BY ?indicationName
    \end{lstlisting}

    \item[CQ6.] \textbf{Which other drugs share the same therapeutic class as \textit{$<$drug x$>$}?}
    \begin{lstlisting}[language=SPARQL, caption={SPARQL query for retrieving drugs with the same therapeutic class}]
        PREFIX mdx: <http://medicX.org/>
    
        SELECT DISTINCT ?otherDrugName ?therapeuticClass
        WHERE {
            mdx:drugX mdx:has_therapeutic_class ?class.
            ?otherDrug mdx:has_therapeutic_class ?class.
            ?otherDrug mdx:name ?otherDrugName.
            FILTER (?otherDrug != mdx:drugX)
        }
        ORDER BY ?otherDrugName
    \end{lstlisting}

    \item[CQ7.] \textbf{Can \textit{$<$drug x$>$} be used during pregnancy or breastfeeding?}
    \begin{lstlisting}[language=SPARQL, caption={SPARQL query for retrieving safety information for drug use in pregnancy or breastfeeding}]
        PREFIX mdx: <http://medicX.org/>

        SELECT DISTINCT ?context ?safetyNote
        WHERE {
            mdx:drugX mdx:has_safety_advisory ?advisory.
            ?advisory mdx:advisory_context ?context ;
                      mdx:safety_note ?safetyNote.
            FILTER (?context IN ("pregnancy", "breastfeeding"))
        }
        ORDER BY ?context
    \end{lstlisting}

\end{enumerate}

The Table~\ref{tab:cq-eval-summary} compares the manually validated answers with those returned by SPARQL queries over the medicX-KG. Small mismatches typically occurred due to query scope (e.g., frequency annotations not always stored as triples).

Note that matching counts between manual entries and SPARQL results do not automatically indicate a "Fully Met" outcome. Evaluation was based not only on the quantity of returned items, but also on their relevance, completeness, and semantic correctness. For example, a CQ may be classified as "Not Met" or "Partially Met" if the returned results lacked clinical specificity, context (e.g., pregnancy vs. breastfeeding), or failed to fully address the intent of the question, even if the item count matched.

\begin{table}[h]
\centering
\caption{Summary of Competency Question (CQ) Evaluation. Each CQ was tested both manually and via SPARQL queries defined above in Appendix~\ref{secA5}. The outcome reflects whether the question could be fully, partially, or not answered using the current version of the KG.}
\label{tab:cq-eval-summary}
\begin{tabular}{|p{7cm}|r|r|c|}
\hline
\rowcolor{gray!25}
\textbf{Competency Question (CQ)} & \textbf{Manual Entries} & \textbf{SPARQL Results}\\
\hline
CQ1: What is the recommended dosage of \textit{Amoxicillin}? & 5 & 5 (Not Met) \\
\hline
\rowcolor{gray!10}CQ2: Which authorised products contain \textit{Paracetamol} in Malta? & 4 & 4 (Fully Met)\\
\hline
CQ3: Are there known interactions between \textit{Warfarin} and \textit{Amlodipine} and \textit{Ativan}? & 6 & 6 (Fully Met)\\
\hline
\rowcolor{gray!10}CQ4: Which adverse drug reactions (side effects) are associated with \textit{Ibuprofen}? & 5 & 4 (Fully Met)\\
\hline
CQ5: For which conditions or diseases is \textit{Metformin} indicated? & 3 & 2 (Partially Met) \\
\hline
\rowcolor{gray!10}CQ6: Which other drugs share the same therapeutic class as \textit{Lisinopril}? & 4 & 4 (Fully Met)\\
\hline
CQ7: Can \textit{Valproate} be used during pregnancy or breastfeeding? & 3 & 3 (Partially Met)\\
\hline
\end{tabular}
\end{table}

Our table with manual–vs–SPARQL comparison is an initial effort. Only seven CQs have been tested several times. CQ3 has been tested with several drugs to verify the ability of KGs to deal with polypharmacy that could lead to DDIs. The answers were judged by two annotators; however, we still require a more thorough evaluation.

\clearpage




\end{appendices}

\pagebreak
\section*{Declarations}

\begin{itemize}
\item \textbf{Ethical approval:} The University Research Ethics Committee (UREC) of the University of Malta provided the approval for this study. Before data collection, all prospective participants received an email that included a letter of consent with all information about the study objectives and the procedure to follow if they agreed to participate. Their agreement to participate and consent to the use of their replies within our research and publications was included in their reply to our email.
\item \textbf{Consent to participate:} Informed consent was obtained from all individual interviewees who participated in our study.
\item \textbf{Consent to publish:} Participants in the interviews consented to their responses to be used in our research and publications.
\item \textbf{Competing interests:} The authors declare that they have no known competing financial interests or personal relationships that could have appeared to influence the work reported in this document.
\item \textbf{Authors' contributions:} L.F. conducted research, development, evaluation of the work, and wrote the draft version of the manuscript text. C.A. supervised the research, contributed to the concept and design, reviewed the draft and wrote the final version of the main text, analysed data, and prepared relevant visualisations. L.M.A contributed to the interview setup and questions, to the analysis of the answers, and reviewed the draft and final version of the document. J.D. contributed to the research and development of the medicX-KG ontology.
\item \textbf{Declaration of Generative AI and AI-assisted technologies in the
writing process:} During the preparation of this work, the authors used ChatGPT to enhance the quality of the English used. After using this tool/service, the authors reviewed and edited the content as needed and assume full responsibility for the content of the publication.
\item \textbf{Funding:} This research was supported by the Malta Digital Innovation Authority (MDIA) through the MAARG 2022 funding scheme.
\item \textbf{Availability of data and materials:} In addition to the data found in the appendices, a sample of the medicX-KG dataset and mapping source code are being made available on github\footnote{\url{https://github.com/charlieabela/medicX-KG}}.
\end{itemize}
\bibliography{sn-bibliography}

\begin{thebibliography}{76}
\providecommand{\natexlab}[1]{#1}
\providecommand{\url}[1]{{#1}}
\providecommand{\urlprefix}{URL }
\providecommand{\doi}[1]{\url{https://doi.org/#1}}
\providecommand{\eprint}[2][]{\url{#2}}
 \bibcommenthead

\bibitem[{{Academy of Managed Care Pharmacy}(2024)}]{amcp2024}
{Academy of Managed Care Pharmacy} (2024) {AMCP Format for Formulary Submissions Version 5.0}. \urlprefix\url{https://www.amcp.org/sites/default/files/2024-04/AMCP-Format-5.0-JMCP-web_0.pdf}

\bibitem[{Asada et~al(2021)Asada, Gunasekaran, Miwa, and Sasaki}]{asada2021}
Asada M, Gunasekaran N, Miwa M, et~al (2021) {Representing a Heterogeneous Pharmaceutical Knowledge-Graph with Textual Information}. Frontiers in Research Metrics and Analytics 6

\bibitem[{Azzopardi and Serracino-Inglott(2020)}]{azzopardi2020}
Azzopardi LM, Serracino-Inglott A (2020) Clinical pharmacy education and practice evolvement in malta. JACCP: Journal Of The American College Of Clinical Pharmacy 3(5):973--979. \doi{https://doi.org/10.1002/jac5.1280}

\bibitem[{Belleau et~al(2008)Belleau, Nolin, Tourigny, Rigault, and Morissette}]{belleau2008}
Belleau F, Nolin MA, Tourigny N, et~al (2008) {Bio2RDF}: towards a mashup to build bioinformatics knowledge systems. Journal of biomedical informatics 41(5):706--716

\bibitem[{Bikaun et~al(2024)Bikaun, Stewart, and Liu}]{bikaun2024}
Bikaun T, Stewart M, Liu W (2024) {CleanGraph: Human-in-the-loop Knowledge Graph Refinement and Completion}. arXiv preprint arXiv:240503932 \urlprefix\url{https://arxiv.org/abs/2405.03932}

\bibitem[{Bodenreider(2004)}]{umls2004}
Bodenreider O (2004) The {Unified Medical Language System} ({UMLS}): integrating biomedical terminology. Nucleic Acids Research 32(Database issue):D267--D270. \doi{10.1093/nar/gkh061}, \urlprefix\url{https://www.nlm.nih.gov/research/umls/}

\bibitem[{Bonner et~al(2022)Bonner, Barrett, Ye, Swiers, Engkvist, Bender, Hoyt, and Hamilton}]{bonner2022}
Bonner S, Barrett IP, Ye C, et~al (2022) A review of biomedical datasets relating to drug discovery: A knowledge graph perspective. Briefings in Bioinformatics 23(6):bbac404

\bibitem[{Braun and Clarke(2006)}]{braun2006}
Braun V, Clarke V (2006) Using thematic analysis in psychology. Qualitative research in psychology 3(2):77--101

\bibitem[{{British Medical Association} and {Royal Pharmaceutical Society}(2024)}]{bnf2024}
{British Medical Association}, {Royal Pharmaceutical Society} (2024) British National Formulary (BNF), 88th edn. BMJ Group and Pharmaceutical Press, London, UK, \urlprefix\url{https://bnf.nice.org.uk/}, accessed April 10, 2025

\bibitem[{Callahan et~al(2020)Callahan, Tripodi, Pielke-Lombardo, and Hunter}]{callahan2020}
Callahan TJ, Tripodi IJ, Pielke-Lombardo H, et~al (2020) Knowledge-based biomedical data science. Annual Review of Biomedical Data Science 3:23--41. \doi{10.1146/annurev-biodatasci-011720-074428}, \urlprefix\url{https://doi.org/10.1146/annurev-biodatasci-011720-074428}

\bibitem[{Chandak et~al(2023)Chandak, Huang, and Zitnik}]{chandak2023}
Chandak P, Huang K, Zitnik M (2023) Building a knowledge graph to enable precision medicine. Scientific Data 10(1):67

\bibitem[{Cook et~al(2023)Cook, Bikkani, and Poterucha~Carter}]{cook2023}
Cook DA, Bikkani A, Poterucha~Carter MJ (2023) Evaluating education innovations rapidly with build-measure-learn: Applying lean startup to health professions education. Medical teacher 45(2):167--178

\bibitem[{Cotovio et~al(2024)Cotovio, Ferraz, Faria, Balbi, Silva, and Pesquita}]{cotovio2024}
Cotovio PG, Ferraz L, Faria D, et~al (2024) {Matcha-DL: A Tool for Supervised Ontology Alignment}. Semantic Web \urlprefix\url{https://www.semantic-web-journal.net/content/matcha-dl-tool-supervised-ontology-alignment}, submitted February 27, 2024; under review

\bibitem[{{Datapharm Ltd}(2024)}]{emc2024}
{Datapharm Ltd} (2024) {Electronic Medicines Compendium (EMC)}. \urlprefix\url{https://www.medicines.org.uk/emc}, accessed: 2024-04-18

\bibitem[{Dennehy et~al(2019)Dennehy, Kasraian, O’Raghallaigh, Conboy, Sammon, and Lynch}]{dennehy2019}
Dennehy D, Kasraian L, O’Raghallaigh P, et~al (2019) A lean start-up approach for developing minimum viable products in an established company. Journal of Decision Systems 28(3):224--232

\bibitem[{{EBSCO Information Services}(2024)}]{dynamed2024}
{EBSCO Information Services} (2024) {DynaMed}. \urlprefix\url{https://www.dynamed.com/}, accessed: 2024-04-18

\bibitem[{{European Commission}(2024)}]{euamr2024}
{European Commission} (2024) Commission staff working document: Analytical document accompanying the {Commission Proposal for a Council Recommendation} on stepping up {EU} actions to combat antimicrobial resistance in a one health approach. Tech. Rep. SWD(2024) 49 final, European Commission, \urlprefix\url{https://health.ec.europa.eu/document/download/085634a9-817d-4f6a-88f2-194ba5f3561e_en?filename=mp_swd_2024_49_en.pdf}, accessed: 2025-04-04

\bibitem[{{European Medicines Agency}(2024)}]{ema2024}
{European Medicines Agency} (2024) {European Medicines Agency: Scientific Evaluation and Supervision of Medicines}. \url{https://www.ema.europa.eu}, accessed April 10, 2025

\bibitem[{Farrugia and Abela(2020)}]{farrugia2020}
Farrugia L, Abela C (2020) {Mining Drug-Drug Interactions for Healthcare Professionals}. In: Proceedings of the 3rd International Conference on Applications of Intelligent Systems. Association for Computing Machinery, New York, NY, USA, APPIS 2020, \doi{10.1145/3378184.3378196}

\bibitem[{Farrugia et~al(2023)Farrugia, Azzopardi, Debattista, and Abela}]{farrugia2023}
Farrugia L, Azzopardi LM, Debattista J, et~al (2023) {Predicting Drug-Drug Interactions Using Knowledge Graphs}. arXiv preprint arXiv:230804172

\bibitem[{Fern{\'a}ndez-Torras et~al(2022)Fern{\'a}ndez-Torras, Duran-Frigola, Bertoni, Locatelli, and Aloy}]{fernandez2022}
Fern{\'a}ndez-Torras A, Duran-Frigola M, Bertoni M, et~al (2022) {Integrating and formatting biomedical data as pre-calculated knowledge graph embeddings in the Bioteque}. Nature Communications 13(1):5304. \doi{10.1038/s41467-022-33026-0}

\bibitem[{{First Databank, Inc.}(2025)}]{fdb2025}
{First Databank, Inc.} (2025) {FDB MedKnowledge Clinical Drug Knowledge}. \url{https://www.fdbhealth.com/}, accessed April 2025

\bibitem[{Flynn et~al(2021)Flynn, Fortier, Maehlen, Pierzinski, and Runnebaum}]{flynn2021}
Flynn AJ, Fortier C, Maehlen H, et~al (2021) A strategic approach to improving pharmacy enterprise automation: Development and initial application of the autonomous pharmacy framework. American Journal of Health-System Pharmacy 78(4):307--316. \doi{10.1093/ajhp/zxaa356}

\bibitem[{Guidotti et~al(2019)Guidotti, Monreale, Ruggieri, Turini, Giannotti, and Pedreschi}]{guidotti2019}
Guidotti R, Monreale A, Ruggieri S, et~al (2019) {A Survey of Methods for Explaining Black Box Models}. ACM Computing Surveys 51(5):93:1--93:42. \doi{10.1145/3236009}

\bibitem[{Hertling and Paulheim(2023)}]{hertling2023}
Hertling S, Paulheim H (2023) {OLaLa}: Ontology matching with large language models. In: Proceedings of the 12th Knowledge Capture Conference 2023, pp 131--139

\bibitem[{Hertling et~al(2022)Hertling, Paulheim, and Shvaiko}]{hertling2022}
Hertling S, Paulheim H, Shvaiko P (2022) Benchmarking ontology matching systems: An updated survey. Applied Sciences 12(13):6606. \doi{10.3390/app12136606}

\bibitem[{Himmelstein et~al(2017)Himmelstein, Lizee, Hessler, Brueggeman, Chen, Hadley, Green, Khankhanian, and Baranzini}]{himmelstein2017}
Himmelstein DS, Lizee A, Hessler C, et~al (2017) Systematic integration of biomedical knowledge prioritizes drugs for repurposing. Elife 6:e26726

\bibitem[{Hou et~al(2023)Hou, Yeung, Xu, Su, Wang, and Zhang}]{hou2023}
Hou Y, Yeung J, Xu H, et~al (2023) {From Answers to Insights: Unveiling the Strengths and Limitations of ChatGPT and Biomedical Knowledge Graphs}. medRxiv \doi{10.1101/2023.07.12.23292499}, \urlprefix\url{https://doi.org/10.1101/2023.07.12.23292499}

\bibitem[{Hyzy et~al(2022)Hyzy, Bond, Mulvenna, Bai, Dix, Leigh, and Hunt}]{hyzy2022}
Hyzy M, Bond R, Mulvenna M, et~al (2022) {System Usability Scale Benchmarking for Digital Health Apps: Meta‑analysis}. JMIR mHealth and uHealth 10(8):e37290. \doi{10.2196/37290}

\bibitem[{{International Pharmaceutical Federation}(2023{\natexlab{a}})}]{fip2023}
{International Pharmaceutical Federation} (2023{\natexlab{a}}) Ethics and the pharmacist: Privacy and confidentiality. \url{https://www.fip.org/file/5591}

\bibitem[{{International Pharmaceutical Federation}(2023{\natexlab{b}})}]{fip2023access}
{International Pharmaceutical Federation} (2023{\natexlab{b}}) Improving access to safe and quality essential medicines and medical devices: The role of pharmacy. \urlprefix\url{https://www.fip.org/file/6065}

\bibitem[{Ioannidis et~al(2020)Ioannidis, Song, Manchanda, Li, Pan, Zheng, Ning, Zeng, and Karypis}]{ioannidis2020}
Ioannidis VN, Song X, Manchanda S, et~al (2020) {DRKG} - drug repurposing knowledge graph for covid-19. \url{https://github.com/gnn4dr/DRKG/}

\bibitem[{Jiménez-Ruiz and Grau(2011)}]{logmap2011}
Jiménez-Ruiz E, Grau BC (2011) {LogMap}: Logic-based and scalable ontology matching. In: The Semantic Web - ISWC 2011, Lecture Notes in Computer Science, vol 7031. Springer, pp 273--288, \doi{10.1007/978-3-642-25073-6_18}

\bibitem[{Keet and {\L}awrynowicz(2016)}]{keet2016}
Keet CM, {\L}awrynowicz A (2016) {Ontology competency questions: A survey, best practices, and proposed CI methodology}. Data \& Knowledge Engineering 105:4--25. \doi{10.1016/j.datak.2016.04.003}

\bibitem[{Knox et~al(2024)Knox, Wilson, Klinger, Franklin, Oler, Wilson, Pon, Cox, Chin, Strawbridge et~al}]{knox2024}
Knox C, Wilson M, Klinger CM, et~al (2024) {DrugBank} 6.0: the drugbank knowledgebase for 2024. Nucleic acids research 52(D1):D1265--D1275

\bibitem[{Koleti et~al(2018)Koleti, Terryn, Stathias, Chung, Cooper, Turner, Vidovic, Forlin, Kelley, D’Urso, Allen, Torre, Jagodnik, Wang, Jenkins, Mader, Niu, Fazel, Mahi, Pilarczyk, Clark, Shamsaei, Meller, Vasiliauskas, Reichard, Medvedovic, Ma’ayan, Pillai, and Sch{\"u}rer}]{koleti2018}
Koleti A, Terryn R, Stathias V, et~al (2018) {Data Portal for the Library of Integrated Network-based Cellular Signatures (LINCS) program: integrated access to diverse large-scale cellular perturbation response data}. Nucleic Acids Research 46(D1):D558--D566. \doi{10.1093/nar/gkx1063}, \urlprefix\url{https://academic.oup.com/nar/article/46/D1/D558/4621325}

\bibitem[{Kontsioti et~al(2022)Kontsioti, Maskell, Bensalem, Dutta, and Pirmohamed}]{kontsioti2022}
Kontsioti E, Maskell S, Bensalem A, et~al (2022) Similarity and consistency assessment of three major online drug-drug interaction resources. Br J Clin Pharmacol 88(9):4067--4079

\bibitem[{Kuhn et~al(2016)Kuhn, Letunic, Jensen, and Bork}]{kuhn2016}
Kuhn M, Letunic I, Jensen LJ, et~al (2016) The {SIDER} database of drugs and side effects. Nucleic acids research 44(D1):D1075--D1079

\bibitem[{Le et~al(2024)Le, Chen, Harris, Fang, Lyn-Cook, Hong, Ge, Rogers, Tong, and Zou}]{le2024}
Le H, Chen R, Harris S, et~al (2024) {RxNorm} for drug name normalization: a case study of prescription opioids in the fda adverse events reporting system. Frontiers in Bioinformatics 3:1328613

\bibitem[{Lipscomb(2000)}]{lipscomb2000}
Lipscomb CE (2000) {Medical Subject Headings (MeSH)}. Bulletin of the Medical Library Association 88(3):265

\bibitem[{Lu and Wang(2025)}]{lu2025}
Lu Y, Wang J (2025) {KARMA: Leveraging Multi-Agent LLMs for Automated Knowledge Graph Enrichment}. arXiv preprint arXiv:250206472 \urlprefix\url{https://arxiv.org/abs/2502.06472}

\bibitem[{{Malta Medicines Authority}(2020)}]{mma2020}
{Malta Medicines Authority} (2020) How medicines are registered. \urlprefix\url{https://medicinesauthority.gov.mt/howmedicinesregistered}, accessed: 2025-04-04

\bibitem[{{Malta Medicines Authority}(2024)}]{mma2024}
{Malta Medicines Authority} (2024) {Malta Medicines Authority}: Regulatory information and services. \url{https://medicinesauthority.gov.mt}, accessed April 10, 2025

\bibitem[{Masnoon et~al(2017)Masnoon, Shakib, Kalisch-Ellett, and Caughey}]{masnoon2017}
Masnoon N, Shakib S, Kalisch-Ellett L, et~al (2017) {What is polypharmacy? A systematic review of definitions}. BMC geriatrics 17:1--10

\bibitem[{Masumshah et~al(2021)Masumshah, Aghdam, and Eslahchi}]{masumshah2021}
Masumshah R, Aghdam R, Eslahchi C (2021) A neural network-based method for polypharmacy side effects prediction. BMC bioinformatics 22:1--17

\bibitem[{{Mayo Clinic}(2023)}]{mayoclinic2023}
{Mayo Clinic} (2023) {Mayo Clinic: Reliable health information and tools for patients}. \url{https://www.mayoclinic.org}, accessed April 2025

\bibitem[{{Medscape, LLC}(2024)}]{medscape2024}
{Medscape, LLC} (2024) {Medscape}. \urlprefix\url{https://www.medscape.com/}, accessed: 2024-04-18

\bibitem[{{Merative}(2024)}]{micromedex2024}
{Merative} (2024) Micromedex solutions. \url{https://www.micromedexsolutions.com/}, accessed April 10, 2025

\bibitem[{Morris et~al(2023)Morris, Soman, Akbas, Zhou, Smith, Meng, Huang, Cerono, Schenk, Rizk-Jackson, Harroud, Sanders, Costes, Bharat, Chakraborty, Pico, Mardirossian, Keiser, Tang, Hardi, Shi, Musen, Israni, Huang, Rose, Nelson, and Baranzini}]{morris2023}
Morris JH, Soman K, Akbas RE, et~al (2023) {The scalable precision medicine open knowledge engine (SPOKE): a massive knowledge graph of biomedical information}. Bioinformatics 39(2):btad080

\bibitem[{Nicholson and Greene(2020)}]{nicholson2020}
Nicholson DN, Greene CS (2020) Constructing knowledge graphs and their biomedical applications. Computational and Structural Biotechnology Journal 18:1414--1428. \doi{10.1016/j.csbj.2020.05.017}, \urlprefix\url{https://doi.org/10.1016/j.csbj.2020.05.017}

\bibitem[{Peng et~al(2023)}]{peng2023}
Peng C, et~al (2023) {Knowledge Graphs: Opportunities and Challenges}. Information 15(8):509. \doi{10.3390/info15080509}

\bibitem[{Percha and Altman(2018)}]{percha2018}
Percha B, Altman RB (2018) {A global network of biomedical relationships derived from text}. Bioinformatics 34(15):2614--2624

\bibitem[{Pi{\~n}ero et~al(2020)Pi{\~n}ero, Ram{\'\i}rez-Anguita, Sa{\"u}ch-Pitarch, Ronzano, Centeno, Sanz, and Furlong}]{pinero2020}
Pi{\~n}ero J, Ram{\'\i}rez-Anguita JM, Sa{\"u}ch-Pitarch J, et~al (2020) The {DisGeNET} knowledge platform for disease genomics: 2019 update. Nucleic acids research 48(D1):D845--D855

\bibitem[{Reese et~al(2021)Reese, Unni, Callahan, Cappelletti, Ravanmehr, Carbon, Shefchek, Good, Balhoff, Fontana et~al}]{reese2021}
Reese JT, Unni D, Callahan TJ, et~al (2021) {KG-COVID-19}: a framework to produce customized knowledge graphs for {COVID-19} response. Patterns 2(1)

\bibitem[{Ren et~al(2022)}]{ren2022}
Ren ZH, et~al (2022) A biomedical knowledge graph-based method for drug–drug interactions prediction through combining local and global features with deep neural networks. Briefings in Bioinformatics 23(5):bbac363. \doi{10.1093/bib/bbac363}

\bibitem[{Robaldo and Batsakis(2024)}]{robaldo2024}
Robaldo L, Batsakis S (2024) On the interplay between validation and inference in {SHACL}: An investigation on the time ontology. Semantic Web 15(3):567--599. \doi{10.3233/SW-240030}, early access, published online 2024‑02‑17

\bibitem[{Romagnoli et~al(2016)Romagnoli, Boyce, Empey, Adams, and Hochheiser}]{romagnoli2016}
Romagnoli KM, Boyce RD, Empey PE, et~al (2016) Bringing clinical pharmacogenomics information to pharmacists: a qualitative study of information needs and resource requirements. International journal of medical informatics 86:54--61

\bibitem[{Rossanez et~al(2020)Rossanez, dos Reis, Torres, and de~Ribaupierre}]{Rossanez2020}
Rossanez A, dos Reis JC, Torres RdS, et~al (2020) {KGen}: a knowledge graph generator from biomedical scientific literature. BMC Medical Informatics and Decision Making 20(4):314

\bibitem[{Sant~Fournier(2000)}]{santfournier2020}
Sant~Fournier MA (2000) Pharmacy and the {EU}: the impact of european union membership on the pharmacy profession in malta

\bibitem[{Scheife et~al(2015)Scheife, Hines, Boyce, Chung, Momper, Sommer, Abernethy, Horn, Sklar, Wong et~al}]{scheife2015}
Scheife RT, Hines LE, Boyce RD, et~al (2015) {Consensus Recommendations for Systematic Evaluation of Drug-Drug Interaction Evidence for Clinical Decision Support}. Drug Safety 38(2):197--206. \doi{10.1007/s40264-014-0262-8}

\bibitem[{Shariff et~al(2022)Shariff, Sridhar, Basha, Alshemeili, and Alzaabi}]{shariff2022}
Shariff A, Sridhar SB, Basha NA, et~al (2022) Development and validation of standardized severity rating scale to assess the consistency of drug-drug interaction severity among various drug information resources. Research in Social and Administrative Pharmacy 18(8):3323--3328

\bibitem[{Shvaiko and Euzenat(2013)}]{shvaiko2013}
Shvaiko P, Euzenat J (2013) Ontology matching: state of the art and future challenges. IEEE Transactions on Knowledge and Data Engineering 25(1):158--176. \doi{10.1109/TKDE.2011.253}

\bibitem[{Sollis et~al(2023)Sollis, Mosaku, Abid, Buniello, Cerezo, Gil, Groza, G{\"u}ne{\c{s}}, Hall, Hayhurst et~al}]{sollis2023}
Sollis E, Mosaku A, Abid A, et~al (2023) The {NHGRI-EBI GWAS} catalog: knowledgebase and deposition resource. Nucleic acids research 51(D1):D977--D985

\bibitem[{Stevenson et~al(2024)Stevenson, Fisher, and D'Souza}]{stevenson2024}
Stevenson R, Fisher G, D'Souza R (2024) The minimum viable product (mvp): Theory and practice. Journal of Management 50(2):123--145. \doi{10.1177/01492063241227154}

\bibitem[{Tan et~al(2023)Tan, Lim, Lim, Ng, and Chng}]{tan2023}
Tan YXF, Lim STY, Lim JL, et~al (2023) Drug information-seeking behaviours of physicians, nurses and pharmacists: A systematic literature review. Health Information \& Libraries Journal 40(2):125--168

\bibitem[{{The UniProt Consortium}(2023)}]{uniprot2023}
{The UniProt Consortium} (2023) Uniprot: the universal protein knowledgebase in 2023. Nucleic Acids Research 51(D1):D523--D531

\bibitem[{{U.S. National Library of Medicine}(2023)}]{medline2023}
{U.S. National Library of Medicine} (2023) {MEDLINE}: Bibliographic database of life sciences and biomedical information. \url{https://www.nlm.nih.gov/medline/}, accessed April 2025

\bibitem[{{U.S. National Library of Medicine}(2025)}]{dailymed2025}
{U.S. National Library of Medicine} (2025) {DailyMed: Current Medication Information}. \url{https://dailymed.nlm.nih.gov/}, accessed April 2025

\bibitem[{Wilkinson et~al(2016)Wilkinson, Dumontier, Aalbersberg, Appleton, Axton, Baak, Blomberg, Boiten, da~Silva~Santos, Bourne et~al}]{wilkinson2016}
Wilkinson MD, Dumontier M, Aalbersberg IJ, et~al (2016) {The FAIR Guiding Principles for scientific data management and stewardship}. Scientific data 3:160018

\bibitem[{Wishart et~al(2023)Wishart, Feunang, Guo, Guo, Sajed, Johnson, Li, Sayeeda, Assempour, Karu, Liu, Lo, Wilson, Knox, Wilson, Feunang, Bernard, Grant, Duggan, Maciejewski, Mahendraker, Zhang, Gautam, Pazanowski, Naseri, Knox, Alam, Golshani, Haider, Wilson, Craig, Eisner, Eisner, Murugesan, Lin, Krishnan, Greiner, Greiner, Williams, Allen, Khan, Ghafourian, Gharagozloo, Alam, Khalek, Parker, Schlessinger, Sinha, Fan, Jones, Joachimiak, Gaulton, Muskal, Liu, McRae, Lyon, Knox, Zheng, McGilvray, Sorokina, Lewis, Law, Wilson, Warner, Nosrati, Noubissi, Choi, Martin, Spicer, Lu, Alfonso, Luis, Hoffman, Magar, Elliott, Ahmed, Currie, Ouellette, Theodoropoulos, Iliopoulos, Yeh, Schmidt, McKenzie, Kothary, Abrahamyan, Khan, Kamal, Rees, White, Ragan, Bateman, and Wishart}]{wishart2023}
Wishart DS, Feunang YD, Guo AC, et~al (2023) {DrugBank} 2023: a comprehensive resource for in silico drug discovery and exploration. Nucleic Acids Research 51(D1):D1144--D1154. \doi{10.1093/nar/gkac1077}, \urlprefix\url{https://go.drugbank.com/}

\bibitem[{{Wolters Kluwer}(2024)}]{uptodate2024}
{Wolters Kluwer} (2024) {UpToDate Clinical Decision Support}. \url{https://www.uptodate.com/}, accessed April 10, 2025

\bibitem[{Xu et~al(2020)Xu, Kim, Song, Jeong, Kim, Kang, Rousseau, Li, Xu, Torvik, Bu, Chen, Ebeid, Li, and Ding}]{Xu2020}
Xu J, Kim S, Song M, et~al (2020) {Building a PubMed knowledge graph}. Scientific Data 7(1):205

\bibitem[{Yan et~al(2024)Yan, Westphal, Seliger, and Usbeck}]{yan2024}
Yan X, Westphal P, Seliger J, et~al (2024) Bridging the gap: Generating a comprehensive biomedical knowledge graph question answering dataset ({PrimeKGQA}). In: Proceedings of the 27th European Conference on Artificial Intelligence (ECAI 2024). IOS Press, pp 1198--1205, \doi{10.3233/FAIA240615}

\bibitem[{Zheng et~al(2021)Zheng, Rao, Song, Zhang, Xiao, Fang, Yang, and Niu}]{zheng2021}
Zheng S, Rao J, Song Y, et~al (2021) {PharmKG}: a dedicated knowledge graph benchmark for bomedical data mining. Briefings in bioinformatics 22(4):bbaa344

\bibitem[{Zhu et~al(2020)Zhu, Nguyen, Grishagin, Southall, Sid, and Pariser}]{zhu2020}
Zhu Q, Nguyen DT, Grishagin I, et~al (2020) An integrative knowledge graph for rare diseases, derived from the genetic and rare diseases information center ({GARD}). Journal of Biomedical Semantics 11(1):1--12. \doi{10.1186/s13326-020-00232-y}

\bibitem[{Zitnik et~al(2018)Zitnik, Agrawal, and Leskovec}]{zitnik2018}
Zitnik M, Agrawal M, Leskovec J (2018) Modeling polypharmacy side effects with graph convolutional networks. Bioinformatics 34(13):i457--i466

\end{thebibliography}

\end{document}